%% file: main.tex
\documentclass[10pt,twocolumn,letterpaper]{article}

\usepackage{cvpr}
\usepackage{times}
\usepackage{epsfig}
\usepackage{graphicx}
\usepackage{amsmath}
\usepackage{amssymb}
\usepackage{makecell}
\usepackage{multirow}
\usepackage{subcaption}
\usepackage[linesnumbered,ruled]{algorithm2e}
\usepackage{pbox}
\usepackage[skip=2pt]{caption}

\usepackage[pagebackref=true,breaklinks=true,letterpaper=true,colorlinks,bookmarks=false]{hyperref}
\usepackage{array}
 
\cvprfinalcopy 

\def\cvprPaperID{967} 

\ifcvprfinal\fi
\begin{document}

\title{SGPN: Similarity Group Proposal Network for 3D  Point Cloud Instance Segmentation}

\author{Weiyue Wang$^1$ \qquad Ronald Yu$^2$
	\qquad Qiangui Huang$^1$  \qquad Ulrich Neumann$^1$\\
	$^1$University of Southern California \hspace{20mm} $^2$University of California, San Diego\\
	\hspace{5mm}Los Angeles, California \hspace{35mm} San Diego, California\\
	\hspace{-10mm}{\tt\small \{weiyuewa,qianguih,uneumann\}@usc.edu}\hspace{20mm}{\tt\small ronaldiscool@gmail.com}\qquad
}

\maketitle

\begin{abstract}
We introduce Similarity Group Proposal Network (SGPN), a simple and intuitive deep learning framework for 3D object instance segmentation on point clouds. 
SGPN uses a single network to  predict point grouping proposals and a corresponding semantic class for each proposal, from which we can directly extract instance segmentation results. Important to the effectiveness of SGPN is its novel representation of 3D instance segmentation results in the form of a similarity matrix that indicates the similarity between each pair of points in embedded feature space, thus producing an accurate grouping proposal for each point. 
Experimental results on various 3D scenes show the effectiveness of our method on 3D instance segmentation, and we also evaluate the capability of SGPN to improve 3D object detection and semantic segmentation results. We also demonstrate its flexibility by seamlessly incorporating 2D CNN features into the framework to boost performance. 
\end{abstract}

\vspace{-3mm}
\input{intro}

\input{relatedworks}

\input{method}

\input{experiments}

\section{Conclusion}
\vspace{-1mm}
We present SGPN, an intuitive, simple, and flexible framework for 3D instance segmentation on point clouds. With the introduction of the similarity matrix as our output representation, group proposals with class predictions can be easily generated from a single network. Experiments show that our algorithm can achieve good performance on instance segmentation for various 3D scenes and  facilitate the tasks of 3D object detection and semantic segmentation. 

\vspace{-3mm}

\paragraph{Future Work} While a similarity matrix provides an intuitive representation and an easily defined loss function, one limitation of SGPN is that the size of the similarity matrix scales quadratically as  $N_p$ increases. Thus, although much more memory efficient than volumetric methods, SGPN cannot process extremely large scenes on the order $10^5$ or more points.
Future research directions can consider generating groups using seeds that are selected based on SGPN to reduce the size of the similarity matrix. 
SGPN can also be extended in future works to learn in a more unsupervised setting or to learn more different kinds of data representations beyond instance segementation.

\appendix
\input{supp}

{\small
\bibliographystyle{ieee}
\bibliography{egbib}
}

\end{document}

%% file: intro.tex
\section{Introduction}
Instance segmentation on 2D images have achieved promising results recently~\cite{He_2017_ICCV,dai2016instance,pinheiro2015learning,li2016fully}. 
With the rise of autonomous driving and robotics applications, the demand for 3D scene understanding and the availability of 3D scene data has rapidly increased in recently. Unfortunately, the literature for 3D instance segmentation and object detection lags far behind its 2D counterpart; scene understanding with Convolutional Neural Networks (CNNs)~\cite{slidingshapes,DeepSlidingShapes,zhuo17amodal3det} on 3D volumetric data is limited by high memory and computation cost.  Recently, deep learning frameworks PointNet/Pointnet++~\cite{pointnet, pointnet2} on point clouds open up more efficient and flexible ways to handle 3D data.

Following the pioneering works in 2D scene understanding, our goal is to develop a novel deep learning framework trained \textit{end-to-end} for \textit{3D instance-aware semantic segmentation on point clouds} that, like established baseline systems for 2D scene understanding tasks, is \textit{intuitive, simple, flexible, and effective}.



\begin{figure}[tb]
\centering
\renewcommand{\arraystretch}{0.1}
\newcolumntype{C}{>{\centering\arraybackslash}p{3.7em}}
\begin{tabular}{CCCCCC}
\multicolumn{3}{c}{\includegraphics[width=0.21\textwidth]{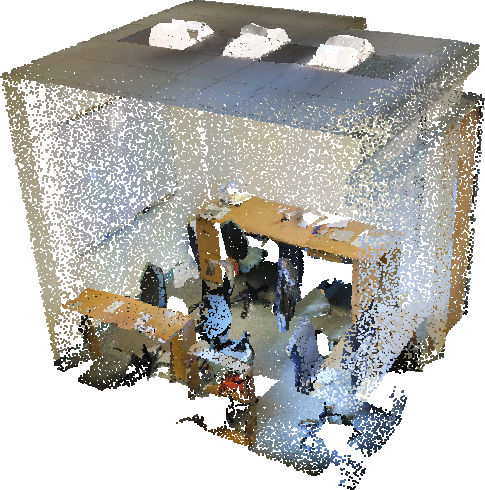}}
&\multicolumn{3}{c}{\includegraphics[width=0.21\textwidth]{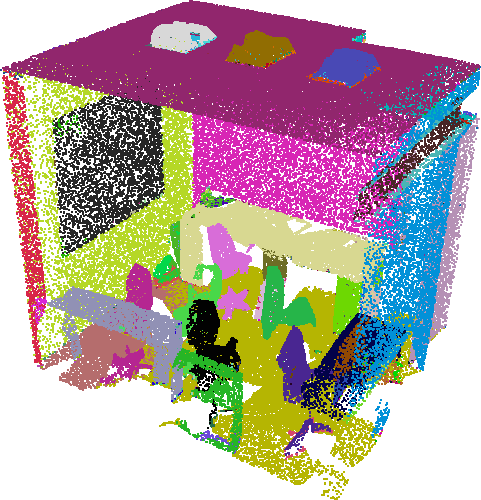}}
\\
\multicolumn{6}{c}{(a)}
\\

\multicolumn{2}{c}{\includegraphics[width=0.08\textwidth]{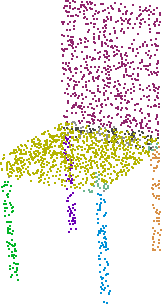}}
&\multicolumn{2}{c}{\includegraphics[width=0.165\textwidth]{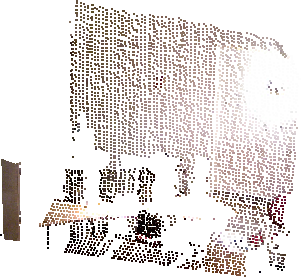}}
&\multicolumn{2}{c}{\includegraphics[width=0.165\textwidth]{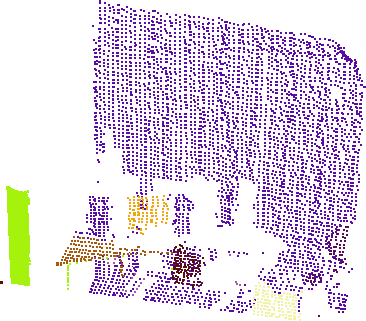}}
\\
\multicolumn{2}{c}{(b)}&\multicolumn{4}{c}{(c)}

\end{tabular}
\caption{Instance segmentation for point clouds using SGPN. Different colors represent different instances. (a) Instance segmentation on complete real scenes. (b) Single object part instance segmentation. (c) Instance segmentation on point clouds obtained from partial scans.}
\label{fig:intro}
\end{figure}

An important consideration for instance segmentation on a point cloud is how to represent output results. Inspired by the trend of predicting proposals for tasks with a variable number of outputs, we introduce a \textit{Similarity Group Proposal Network (SGPN)}, which formulates \textit{group proposals} of object instances by learning a novel 3D instance segmentation representation in the form of a \textit{similarity matrix} .

Our pipeline first uses PointNet/PointNet++ to extract a descriptive feature vector \textit{for each point} in the point cloud.
As a form of similarity metric learning, we enforce the idea that points belonging to the same object instance should have very similar features; hence we measure the distance between the features of each pair of points in order to form a \textit{similarity matrix} that indicates whether any given pair of points belong to the same object instance.

The rows in our similarity matrix can be viewed as instance candidates, which we combine with learned confidence scores in order to generate plausible \textit{group proposals}.
We also learn a semantic segmentation map in order to classify each object instance obtained from our group proposals.
We are also able to directly derive tight 3D bounding boxes for object detection.

By simply measuring the distance between overdetermined feature representations of each pair of points, our \textit{similarity matrix} simplifies 
our pipeline in that we remain in the natural point cloud representation of defining our objects by the relationships between points.


In summary, SGPN has three output branches for instance segmentation on point clouds: a \textit{similarity matrix} yielding point-wise group proposals, a \textit{confidence map} for pruning these proposals, and a \textit{semantic segmentation map} to give the class label for each group.

We evaluate our framework on both 3D shapes (ShapeNet~\cite{shapenet}) and real 3D scenes (Stanford Indoor Semantic Dataset~\cite{armeni_cvpr16} and NYUV2~\cite{silberman2012indoor}) and demonstrate that SGPN achieves state-of-the-art results on 3D instance segmentation. We also conduct comprehensive experiments to show the capability of SGPN on achieving high performance on 3D semantic segmentation and 3D object detection on point clouds.
Although a minimalistic framework with no bells and whistles already gives visually pleasing results
({Figure~\ref{fig:intro}}), we also demonstrate the flexibility of SGPN as we boost performance even more by seamlessly integrating CNN features from RGBD images.



%% file: relatedworks.tex
\section{Related Works}

\subsection{Object Detection and Instance Segmentation}\label{sec:detection}
Recent advances in object detection ~\cite{ren2015faster,girshick2015fast, Lin_2017_CVPR, redmon2016you,Redmon_2017_CVPR, liu2016ssd,fu2017dssd,lin2017focal} and instance segmentation ~\cite{li2016fully, dai2016instance, dai2016instance2, pinheiro2016learning, pinheiro2015learning} on 2D images have achieved promising results. R-CNN ~\cite{girshick14CVPR} for 2D object detection established a baseline system by introducing region proposals as candidate object regions. Faster R-CNN ~\cite{ren2015faster} leveraged a CNN learning scheme and proposed Region Proposal Networks(RPN). YOLO~\cite{redmon2016you} divided the image into grids and each grid cell produced an object proposal. Many 2D instance segmentation approaches are based on segment proposals. DeepMask~\cite{pinheiro2015learning} learns to generate segment proposals each with a corresponding object score. Dai et al.~\cite{dai2016instance} predict segment candidates from bounding box proposals. Mask R-CNN ~\cite{He_2017_ICCV} extended Faster R-CNN by adding a branch on top of RPN to produce object masks for instance segmentation.

Following these pioneering 2D works, 3D bounding box detection frameworks have emerged ~\cite{ren2016three, slidingshapes, DeepSlidingShapes, zhuo17amodal3det, Chen_2017_CVPR}. Song and Xiao~\cite{DeepSlidingShapes} use a volumetric CNN to create 3D RPN on a voxelized 3D scene and then use both the color and depth data of the image in a joint  3D and 2D object recognition network on each proposal.
Deng and Latecki ~\cite{zhuo17amodal3det}  regress
class-wise 3D bounding box models based on RGBD image
appearance features only.
Armeni et al~\cite{armeni_cvpr16} use a sliding shape method with CRF to perform 3D object detection on point cloud.
To the best of our knowledge, no previous work exists that learns 3D instance segmentation.

\subsection{3D Deep Learning} \label{3dlearning}

Convolutional neural networks generalize well to 3D by performing convolution on voxels for certain tasks such as object classification ~\cite{qi2016volumetric, wang2017cnn, maturana2015voxnet, wu20153d, Riegler2017OctNet,guan2016icpr,guan2015mva},
shape reconstruction ~\cite{Wang_2017_ICCV, HAN_2017_ICCV,  dai2017complete} of simple objects, and 3D object detection as mentioned in Section~\ref{sec:detection}. However, volumetric representation carry a high memory and computational cost and have strong limitations dealing with 3D scenes~\cite{scannet,armeni_cvpr16,song2016ssc}. Octree-based CNNs~\cite{Riegler2017OctNet, ogn2017, wang2017cnn} have been introduced recently, but they are less flexible than volumetric CNNs and still suffer from memory efficiency problems.


A point cloud is an intuitive, memory-efficient 3D representation well-suited for representing detailed, large scenes for 3D instance segmentation using deep learning.
 PointNet/Pointnet++~\cite{pointnet, pointnet2} recently introduce deep neural networks on 3D point clouds, learning successful results for tasks such as object classification and part and semantic scene segmentation. We base our network architecture off of PointNet/PointNet++, achieving a novel method that learns 3D instance segmentation on point clouds.

\subsection{Similarity Metric Learning}
Our work is also closely related to similarity metric learning, which has been widely used in deep learning on various tasks such as person re-identification~\cite{deepmetric_person}, matching~\cite{matchnet_cvpr_15}, image retrival~\cite{frome2007learning,weinberger2009distance} and face recognition~\cite{lecun_cvpr_05}. Siamese CNNs~\cite{lecun_cvpr_05,simo2015discriminative, bertinetto2016fully} are used on tasks such as tracking~\cite{lealcvprw2016} and one-shot learning~\cite{koch2015siamese} by measuring the similarity of two input images. Alejandro et. al~\cite{Alejandro2017nips} introduced an associative embedding method to group similar pixels for multi-person pose estimation and 2D instance segmentation by enforcing that
 pixels in the same group should have similar values in their embedding space without actually enforcing what those exact values should be.
  Our method exploits metric learning in a different way in that we regress the likelihood of two points belonging to the same group and formulate the similarity matrix as group proposals to handle variable number of instances.

%% file: method.tex
\section{Method}
\begin{figure*}[tb]
\centering
\includegraphics[width=0.95\textwidth]{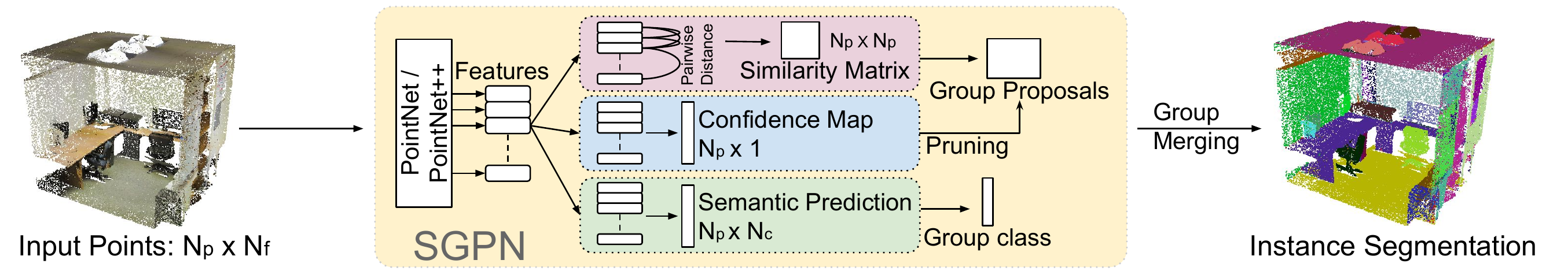}
\caption{Pipeline of our system for point cloud instance segmentation.}
\vspace{-3mm}
\label{fig:overview}
\end{figure*}

The goal of this paper is to take a 3D point cloud as input and produce an object instance label for each point and a class label for each instance. Utilizing recent developments in deep learning on point clouds~\cite{pointnet,pointnet2}, we introduce a Similarity Group Proposal Network (SGPN), which  consumes a 3D point cloud and outputs a set of instance proposals that each contain the group of points inside the instance as well as its class label. 
Section~\ref{sec:sgpn} introduces the design and properties of SGPN. Section~\ref{sec:grouping} proposes an algorithm  to merge similar groups and give each point an instance label.  Section~\ref{sec:implement} gives implementation details. Figure~\ref{fig:overview} depicts the overview of our system.

\subsection{Similarity Group Proposal Network}
\label{sec:sgpn}

SGPN is a very \textit{simple} and \textit{intuitive} framework. As shown in Figure \ref{fig:overview}, it first passes a point cloud $P$ of size $N_p$ through a feed-forward feature extraction network inspired by PointNets~\cite{pointnet,pointnet2}, learning both global and local features in the point cloud.
This feature extraction network produces a matrix $F$. 
SGPN then diverges into three branches that each pass $F$ through a single PointNet layer to obtain
sized $N_p\times N_f$ feature matrices $F_{SIM}, F_{CF}, F_{SEM}$, which we respectively use to obtain a \textit{similarity matrix}, a \textit{confidence map} and a \textit{semantic segmentation map}. 
The $i$th row in a $N_p\times N_f$ feature matrix is a $N_f$-dimensional vector that represents point $P_i$ in an embedded feature space.
Our loss $L$ is given by the sum of the losses from each of these three branches: $L=L_{SIM}+L_{CF}+L_{SEM}$. Our network architecture can be found in the supplemental.

\paragraph{Similarity Matrix}
\label{sgpm}
We propose a novel \textit{similarity matrix} $S$ from which we can formulate group proposals to directly recover accurate instance segmentation results.
$S$ is of dimensions $N_p\times N_p$, and element $S_{ij}$ classifies whether or not points $P_i$ and $P_j$ belong to the same object instance. Each row of $S$ can be viewed as a proposed grouping of points that form a candidate object instance.

We leverage that points belonging to the same object instance should have similar features and lie very close together in feature space. We obtain $S$ by, for each pair of points $\{P_i,P_j\}$, simply subtracting their corresponding feature vectors $\{F_{SIM_i}$, $F_{SIM_j}\}$ and taking the $L_2$ norm such that $S_{ij}=||F_{SIM_i}-F_{SIM_j}||_2$.
This reduces the problem of instance segmentation to learning an embedding space where points in the same instance are close together and those in different object instances are far apart. 

\begin{figure}[tb]
\centering
\renewcommand{\arraystretch}{0.1}
\begin{tabular} {p{2.5em}p{2.5em}p{2.5em}|p{2.5em}p{2.5em}p{2.5em}}
\includegraphics[width=0.05\textwidth]{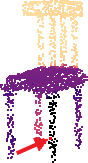}
&\includegraphics[width=0.05\textwidth]{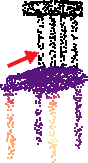}
&\includegraphics[width=0.06\textwidth]{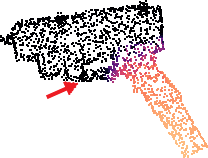}
&
\raisebox{.45\height}{\includegraphics[width=0.08\textwidth]{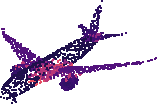}}
&\includegraphics[width=0.04\textwidth]{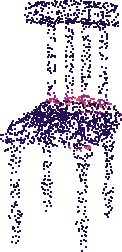}
&\includegraphics[width=0.04\textwidth]{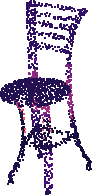}
\\
\includegraphics[width=0.07\textwidth]{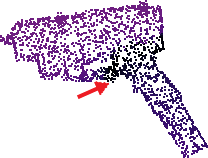}
&\includegraphics[width=0.06\textwidth]{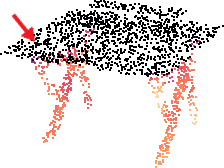}
&\includegraphics[width=0.06\textwidth]{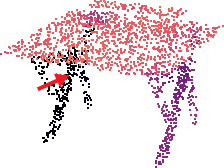}
&
\includegraphics[width=0.06\textwidth]{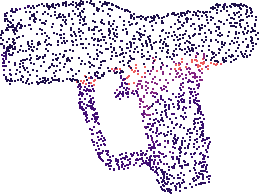}
&\includegraphics[width=0.06\textwidth]{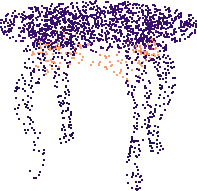}
&\raisebox{.2\height}{
\includegraphics[width=0.07\textwidth]{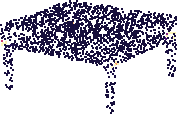}}
\\
&(a)& & &(b)&
\end{tabular}
\caption{(a) Similarity (euclidean distance in feature space) between a given point (indicated by red arrow) and the rest of points.  A darker color represents lower distance in feature space thus higher similarity. (b) Confidence map. A darker color represents higher confidence.}
\vspace{-3mm}
\label{fig:ptssimilarity}
\end{figure}

For a better understanding of how SGPN captures correlation between points, in Figure~\ref{fig:ptssimilarity}(a) we visualize the similarity (euclidean distance in feature space) between a given point and the rest of the points in the point cloud. Points in different instances have greater euclidean distances in feature space and thus smaller similarities even though they have the same semantic labels. For example, in the bottom-right image of Figure~\ref{fig:ptssimilarity}(a), although the given table leg point has greater similarity with the other table leg points than the table top, it is still distinguishable from the other table leg.

We believe that a \textit{similarity matrix} is a more natural and simple representation for 3D instance segmentation on a point cloud compared to traditional 2D instance segmentation representations.
Most state-of-the-art 2D deep learning methods for instance segmentation first localize the image into patches,
which are then passed
through a neural network and segment a binary mask of the object.

While learning a binary mask in a bounding box is a more natural representation for  \textit{space-centric} structures such as images or volumetric grids where features are largely defined by which positions in a grid have strong signals, point clouds can be viewed as \textit{shape-centric} structures where information is encoded by the relationship between the points in the cloud, so we would prefer to also define instance segmentation output by the relationship between points without working too much in grid space.


Hence we expect that a deep neural network could better learn our \textit{similarity matrix}, which compared to traditional representations is a \textit{more natural and straightforward} representation for instance segmentation in a point cloud.

\paragraph{Double-Hinge Loss for Similarity Matrix} As is the case in ~\cite{Alejandro2017nips}, in our similarity matrix we do not need to precisely regress the exact values of our features; we only optimize the simpler objective that similar points should be close together in feature space. We define three potential similarity classes for each pair of points $\{P_i,P_j\}$: 1) $P_i$ and $P_j$ belong to the same object instance;
2) $P_i$ and $P_j$ share the same semantic class but do not belong to the same object instance;
3) $P_i$ and $P_j$ do not share the same semantic class.
Pairs of points should lie progressively further away from each other in feature space as their similarity class increases.
We define out loss as:
\begin{gather*}
L_{SIM}=\sum_{i}^{N_p}\sum_{j}^{N_p} l(i,j)\\
l(i,j)=
\begin{cases}
 ||F_{SIM_i}-F_{SIM_j}||_2 & C_{ij}=1\\
\alpha\max(0,K_1-||F_{SIM_i}-F_{SIM_j}||_2) & C_{ij}=2\\
\max(0,K_2-||F_{SIM_i}-F_{SIM_j}||_2) & C_{ij}=3
\end{cases}
\end{gather*}
where $C_{ij}$ indicates which of the similarity classes defined above does the pair of points $(\{P_i, P_j)\}$ belong to and $\alpha,K_1,K_2$ are constants such that $\alpha>1$, $K_2>K_1$.

Although the second and third similarity class are treated equivalently for the purposes of instance segmentation, distinguishing between them in $L_{SIM}$ using our double-hinge loss allows our similarity matrix output branch and our semantic segmentation output branch to mutually assist each other for increased accuracy and convergence speed. Since the semantic segmentation network is actually wrongly trying to bring pairs of points in our second similarity class closer together in feature space, we also add an $\alpha>1$ term to increase the weight of our loss to dominate the gradient from the semantic segmentation output branch.

At test time if $S_{ij} < Th_S$ where  $Th_S<K_1$, then points pair $P_i$ and $P_j$ are in the same instance group.

\paragraph{Similarity Confidence Map}
SGPN also feeds $F_{CF}$ through an additional PointNet layer to predict a $N_p\times 1$ confidence map $CM$ reflecting how confidently the model believes that each grouping candidate is indeed a correct object instance. Figure~\ref{fig:ptssimilarity}(b) provides a visualization of the confidence map; points located in the boundary area between parts have lower confidence.

We regress confidence scores based on ground truth groups $G$ represented as a $N_p\times N_p$ matrix identical in form to our similarity matrix. If $P_i$ is a background point that does not belong to any object in the ground truth then the row $G_i$ will be all zeros.
For each row in $S_i$, we expect the ground-truth value in the confidence map $CM_i$ to be the intersection over union (IoU) between the set of points in the predicted group $S_i$ and the ground truth group $G_i$. Our loss $L_{CF}$ is the L2 loss between the inferred and expected $CM$.

Although the loss $L_{CF}$ depends on the similarity matrix output branch during training, at test time we run the branches in parallel and only groups with confidence greater than a threshold $Th_C$  are considered valid group proposals.

\paragraph{Semantic Segmentation Network}
The semantic segmentation map acts as a point-wise classifier. 
SGPN passes $F_{SEM}$ through an additional PointNet layer whose architecture depends on the number of possible semantic classes, yielding the final output $M_{SEM}$, which is a $N_p\times N_C$ sized matrix where $N_C$ is the number of possible object categories. 
$M_{SEM_{ij}}$ corresponds to the probability that point $P_i$ belongs to class $C_j$.

The loss $L_{SEM}$ is a weighted sum of the cross entropy softmax loss for each row in the matrix.
We use median frequency balancing~\cite{badrinarayanan2015segnet} and the weight assigned to a category is $ac = median freq/freq(c)$, where $freq(c)$ is the total number of points of class $c$ divided by the total number of points in samples where $c$ is present, and $median freq$ is the median of these $freq(c)$.

At test time, the class label for a group instance is assigned by calculating the mode of the semantic labels of the points in that group.

\subsection{Group Proposal Merging}
\label{sec:grouping}
The similarity matrix $S$ produces $N_p$ group proposals, many of which are noisy or represent the same object. We first discard proposals with predicted confidence less than  $Th_C$ or cardinality less than $Th_{M2}$. We further prune our proposals into clean, non-overlapping object instances by applying Non-Maximum Suppression; groups with IoU greater than $Th_{M1}$ are merged together by selecting the group with the maximum cardinality. 

Each point is then assigned to the group proposal that contains it. In the rare case ($\sim 2\%$) that after the merging stage a point belongs to more than one final group proposal, this usually means that the point is at the boundary between two object instances, which means that the effectiveness of our network would be roughly the same regardless of which group proposal the point is assigned to.  Hence, with minimal loss in accuracy we randomly assign the point to any one of the group proposals that contains it. We refer to this process as \textit{GroupMerging} throughout the rest of the paper. 

%

\subsection{Implementation Details}
\label{sec:implement}

We use an ADAM~\cite{kingma2014adam} optimizer with initial learning rate $0.0005$, momentum $0.9$ and batch size $4$. The learning rate is divided by $2$ every $20$ epochs. The network is trained with only the $L_{SIM}$ loss for the first $5$ epochs. In our experiment, $\alpha$ is set to $2$ initially and is increased by $2$ every $5$ epochs. This design makes the network more focused on separating features of points that belong to different object instances but have the same semantic labels. $K_1, K_2$ are set to $1.0$ and $2.0$, respectively. We use per-category histogram thresholding to get the threshold point $Th_s$ for each testing sample. $Th_{M1}$ is set to $0.6$ and $Th_{M2}$ is set to $200$. $Th_C$ is set to $0.1$.
Our network is implemented with Tensorflow and a single Nvidia GTX1080 Ti GPU. It takes 16-17 hours to converge. At test time, SGPN takes $40$ms on an input point cloud with size $4096\times 9$ with PointNet++ as our baseline architecture. Further runtime analysis can be found in Section~\ref{nyu}. Code is availabel at \url{github.com/laughtervv/SGPN}.

%% file: experiments.tex
\section{Experiments}

We evaluate SGPN on 3D instance segmentation on the following datasets:
\begin{itemize}
\item Stanford 3D Indoor Semantics Dataset (S3DIS)~\cite{armeni_cvpr16}: This dataset contains 3D scans in $6$ areas including $271$ rooms. The input is a complete point cloud generated from scans fused together from multiple views. Each point has semantic labels and instance annotations.
\item NYUV2~\cite{silberman2012indoor}: Partial point clouds are generated from single view RGBD images. The dataset is annotated with 3D bounding boxes and 2D semantic segmentation masks. We use the improved annotation in ~\cite{zhuo17amodal3det}. Since both 3D bounding boxes and 2D segmentation masks annotations are given, ground truth 3D instance segmentation labels for point clouds can be easily generated 
We follow the standard split with 795 training images and 654 testing images.
\item ShapeNet~\cite{shapenet,yi2016scalable} Part Segmentation: ShapeNet contains $16,881$ shapes annotated with $50$ types of parts in total. Most object categories are labeled with two to five parts. We use the official split of 795 training samples and 654 testinn percentageg samples in our experiments.
\end{itemize}
We also show the capability of SGPN to improve semantic segmentation and 3D object detection. To validate the flexibility of SGPN, we also seamlessly incorporate 2D CNN features into our network to boos performance on the NYUV2 dataset.

\subsection{S3DIS Instance Segmentation and 3D Object Detection}
We perform experiments on Stanford 3D Indoor Semantic Dataset to evaluate our performance on large real scene scans. Following experimental settings in PointNet~\cite{pointnet}, points are uniformly sampled into blocks of area $1m \times 1m$. Each point is labeled as one of $13$ categories (chair, table, floor, wall, clutter etc.) and represented by a $9$D vector (XYZ, RGB, and normalized location as to the room). At train time we uniformly sample $4096$ points in each block, and  at test time we use all points in the block as input. 


SGPN uses PointNet as its baseline architecture for this experiment.\footnote{PointNet~\cite{pointnet} proposed a 3D detection system while PointNet++~\cite{pointnet2} does not. To make fair comparison, we use PointNet as our baseline architecture for this experiment while using PointNet++ in Sections~\ref{nyu} and \ref{sec:experimentshapenet}.}
Figure~\ref{fig:S3DISins} shows instance segmentation results on S3DIS with SGPN. Different colors represent different instances. Point colors of the same group are not necessarily the same as their counterparts in the ground truth since object instances are unordered. To visualize instance classes, we also add semantic segmentation results. SGPN achieves good performance on various room types.

We also compare instance segmentation performance with the following method (which we call Seg-Cluster): Perform semantic segmentation using our network and then select all points as seeds. Starting from a seed point, BFS is used to search neighboring points with the same label. If a cluster with more than $200$ points has been found, it is viewed as a valid group. Our \textit{GroupMerging} algorithm is then used to merge these valid groups.

\begin{table*}
\begin{center}
\setlength{\tabcolsep}{0.21em}
\newcolumntype{C}{>{\centering\arraybackslash}p{2.97em}}
\begin{tabular}{c|C|CCCCCCCCCCCC}
\Xhline{3\arrayrulewidth}
& \small Mean& ceiling& floor& wall& beam& column& window& door& table& chair& sofa& bookcase& board\\
\hline
Seg-Cluster& 17.40 & \bf{70.01} &80.12& 10.64& 15.30& 0.00& 28.97& 32.32& 22.16& 27.76& 0.00& 0.06& 21.52\\
SGPN & \bf{36.30} &  58.42 & \bf{83.67} & \bf{42.24} & \bf{25.64} & \bf{7.15} & \bf{42.73}& \bf{45.23}& \bf{38.25}& \bf{47.05}&0.00& \bf{13.57}&  \bf{31.68}\\

  \Xhline{3\arrayrulewidth}
\end{tabular}
\end{center}
\vspace{-3mm}
\caption{Results on instance segmentation in S3DIS scenes. The metric is AP(\%) with IoU threshold $0.5$. To the best of our knowledge, there are no existing instance segmentation methods on point clouds for arbitrary object categories.}
\vspace{-4mm}
\label{table:S3DISinstance}
\end{table*}

\begin{table}
\begin{center}
\setlength{\tabcolsep}{0.1em}
\newcolumntype{C}{>{\centering\arraybackslash}p{2.4em}}
\begin{tabular}{c|c|c|c}
\Xhline{3\arrayrulewidth}
& \small $AP_{0.25}$ & $AP_{0.5}$ & $AP_{0.75}$\\
\hline
Seg-Cluster&  34.8 & 17.4 & 11.2 \\
SGPN    & \bf{52.6} & \bf{36.3} & \bf{18.8} \\
  \Xhline{3\arrayrulewidth}
\end{tabular}
\end{center}
\caption{Comparison results on instance segmentation with different IoU thresholds in S3DIS scenes. Metric is mean AP(\%) over $13$ categories.}
\label{table:differentious}
\end{table}

\begin{table}
\begin{center}
\setlength{\tabcolsep}{0.1em}
\newcolumntype{C}{>{\centering\arraybackslash}p{2.4em}}
\begin{tabular}{c|C|CCCC}
\Xhline{3\arrayrulewidth}
& \small Mean& table & chair & sofa & board\\
\hline
PointNet~\cite{pointnet}& 24.24 &46.67 & 33.80 & 4.76 & 11.72\\
Seg-Cluster& 18.72 & 33.44 & 22.8 & 5.38 & 13.07\\
SGPN    & \bf{30.20}   & \bf{49.90} & \bf{40.87} &\bf{6.96} & \bf{13.28} \\
  \Xhline{3\arrayrulewidth}
\end{tabular}
\end{center}
\vspace{-3mm}
\caption{Comparison results on 3D detection in S3DIS scenes. SGPN uses PointNet as baseline. The metric is AP with IoU threshold $0.5$.}
\vspace{-2mm}
\label{table:S3DISdetection}
\end{table}

\begin{table}
\begin{center}
\setlength{\tabcolsep}{0.1em}
\newcolumntype{C}{>{\centering\arraybackslash}p{2.4em}}
\begin{tabular}{c|c|c}
\Xhline{3\arrayrulewidth}
& \small Mean IoU& Accuracy\\
\hline
PointNet~\cite{pointnet}& 49.76 & 79.66\\
SGPN & \bf{50.37}& \bf{80.78} \\
  \Xhline{3\arrayrulewidth}
\end{tabular}
\end{center}
\vspace{-3mm}
\caption{Results on semantic segmentation in S3DIS scenes. SGPN uses PointNet as baseline. Metric is mean IoU(\%) over $13$ classes (including clutter).}
\vspace{-2mm}
\label{table:S3DISsemseg}
\end{table}

We calculate the IoU on points between each predicted and ground truth group. A detected instance is considered as true positive if the IoU score is greater than a threshold. The average precision (AP) is further calculated for instance segmentation performance evaluation. Table ~\ref{table:S3DISinstance} shows the AP for every category with IoU threshold $0.5$. To the best of our knowledge, there are no existing instance segmentation method on point clouds for arbitrary object categories, so we further demonstrate the capability of SGPN to handle various objects by adding the 3D detection results of Armeni et al. ~\cite{armeni_cvpr16} on S3DIS to Table ~\ref{table:S3DISinstance}. The difference in evaluation metrics between our method and ~\cite{armeni_cvpr16} is that the IoU threshold of ~\cite{armeni_cvpr16} is $0.5$ on a 3D bounding box and the IoU calculation of our method is on points. Despite this difference in metrics, we can still see our superior performance on both large and small objects.

\begin{figure}[tb]
\centering
\newcolumntype{C}{>{\centering\arraybackslash}p{3.6em}}
\begin{tabular}{CCCCC}
\includegraphics[width=0.093\textwidth]{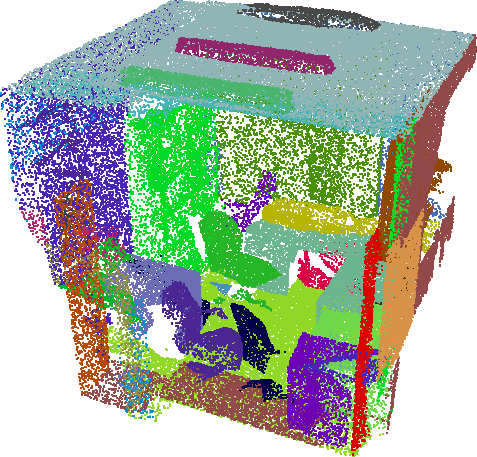}
&\includegraphics[width=0.093\textwidth]{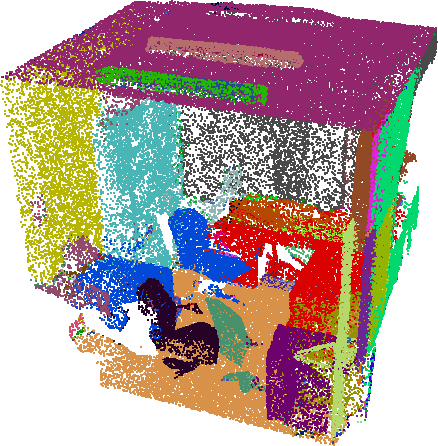}
&\includegraphics[width=0.093\textwidth]{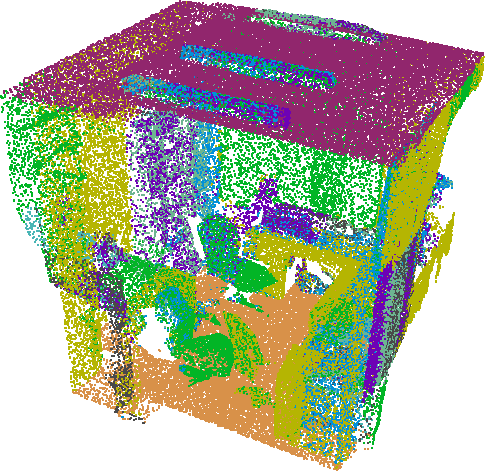}
&\includegraphics[width=0.093\textwidth]{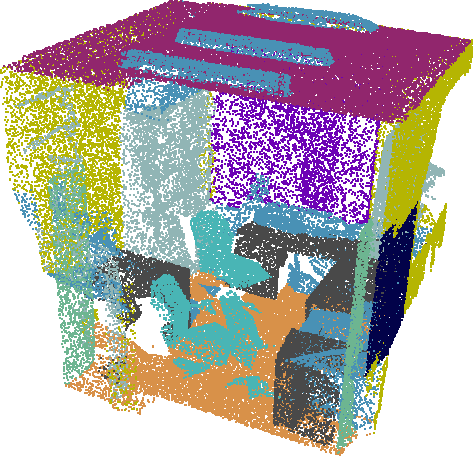}
&\includegraphics[width=0.093\textwidth]{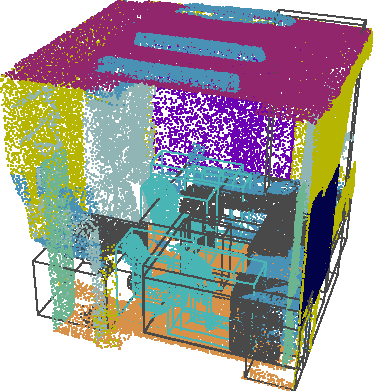}
\\
\includegraphics[width=0.093\textwidth]{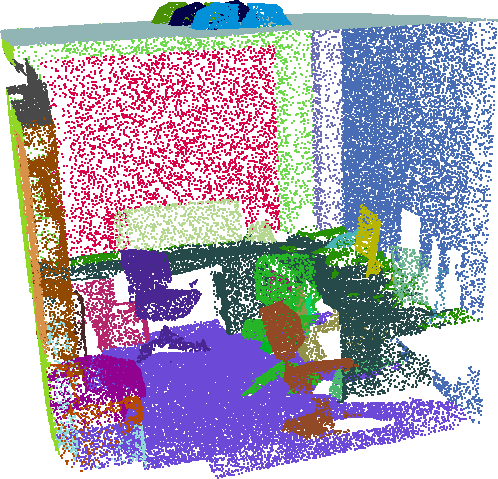}
&\includegraphics[width=0.093\textwidth]{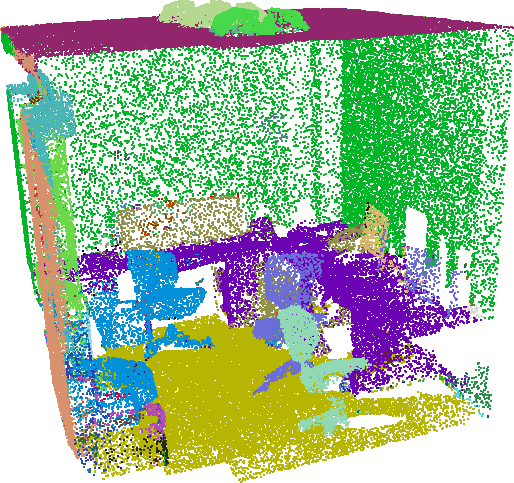}
&\includegraphics[width=0.093\textwidth]{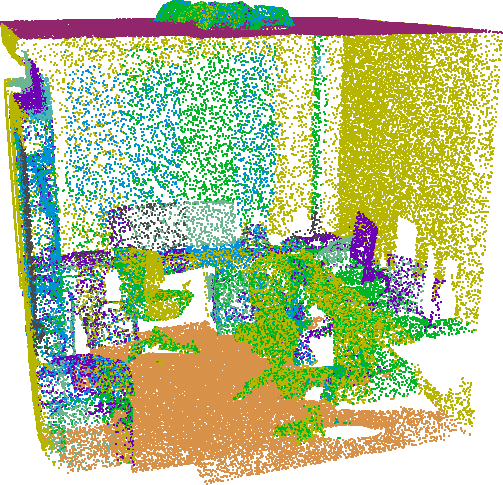}
&\includegraphics[width=0.093\textwidth]{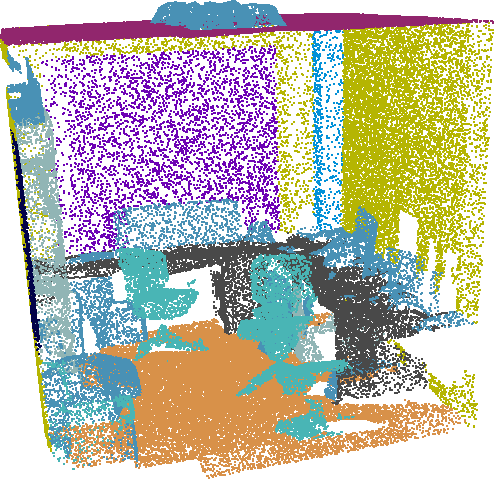}
&\includegraphics[width=0.093\textwidth]{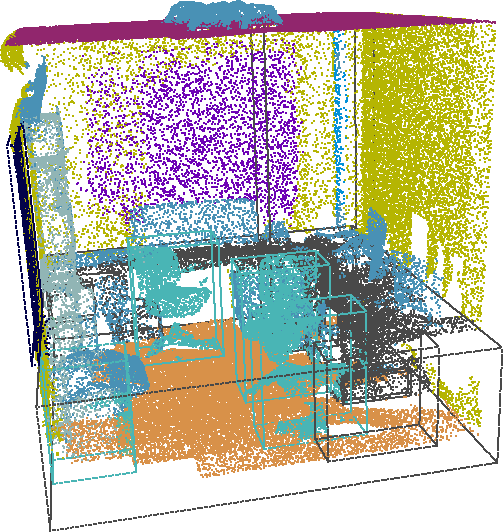}
\\
\includegraphics[width=0.093\textwidth]{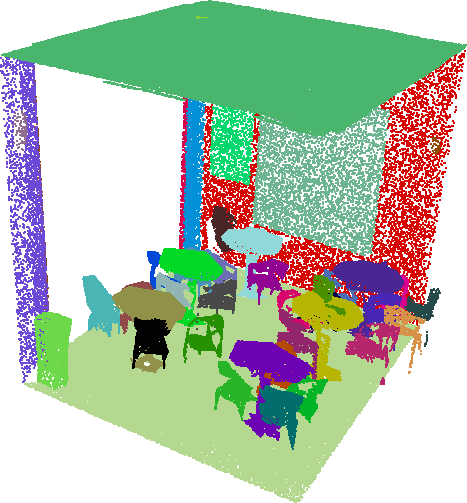}
&\includegraphics[width=0.093\textwidth]{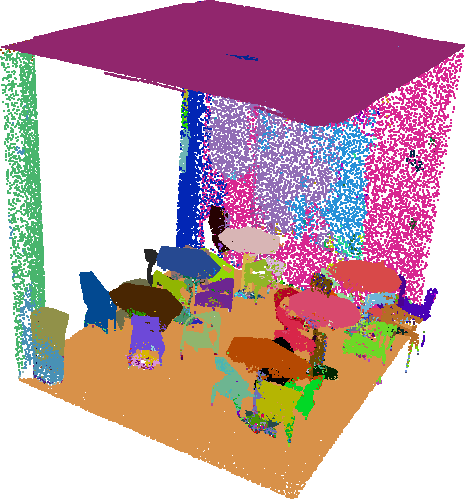}
&\includegraphics[width=0.093\textwidth]{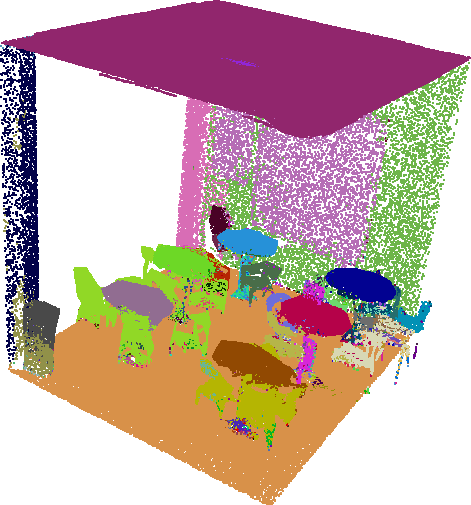}
&\includegraphics[width=0.093\textwidth]{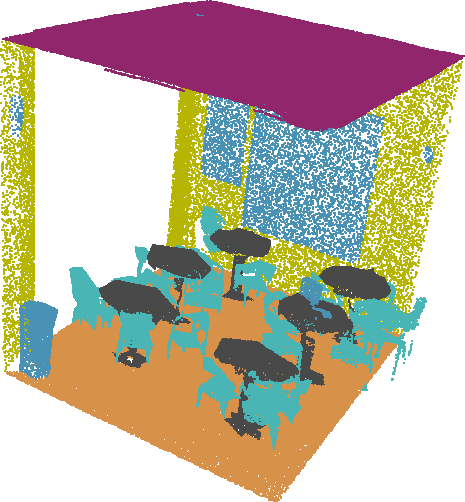}
&\includegraphics[width=0.093\textwidth]{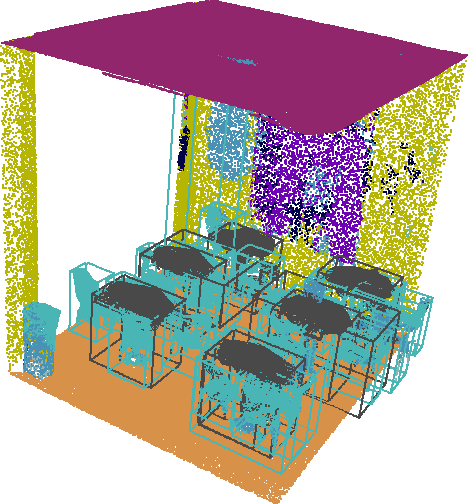}
\\
(a) & (b) & (c) & (d) & (e)
\end{tabular}
\caption{Comparison results on S3DIS. (a) Ground Truth for instance segmentation. Different colors represents different instances. (b) SGPN instance segmentation results. (c) Seg-Cluster instance segmentation results. (d) Ground Truth for semantic segmentation. (e) Semantic Segmentation and 3D detection results of SGPN. The color of the detected bounding box for each object category is the same as the semantic labels.}
\label{fig:S3DISres}
\end{figure}

We see that a naive method like Seg-Cluster tends to properly separate regions far away for large objects like the ceiling and floor. However for small object, Seg-Cluster fails to segment instances with the same label if they are close to each other. Mean APs with different IoU thresholds ($0.25$, $0.5$, $0.75$) are also evaluated in Table~\ref{table:differentious}.
Figure~\ref{fig:S3DISres} shows qualitative comparison results.

Once we have instance segmentation results, we can compute the bounding box for every instance and thus produce 3D object detection predictions. In Table \ref{table:S3DISdetection}, we compare out method with the 3D object detection system introduced in PointNet ~\cite{pointnet}, which to the best of our knowledge is the state-of-the-art method for 3D detection on S3DIS. Detection performance is evaluated over $4$ categories AP with IoU threshold $0.5$.

The method introduced in PointNet clusters points given semantic segmentation results and uses a binary classification network for each category to separate close objects with same categories. 
Our method outperforms it by a large margin, and unlike PointNet does not require an additional network, which unnecessarily introduces additional complexity during both train and test time and local minima during train time. SGPN can effectively separate the difficult cases of objects of the same semantic class but different instances (c.f. Figure~\ref{fig:S3DISres}) since points in different instances are far apart in feature space even though they have the same semantic label. We further compare our semantic segmentation results with PointNet in Table~\ref{table:S3DISsemseg}. SGPN outperforms its baseline with the help of its similarity matrix.

\begin{figure}[tb]
\centering
\newcolumntype{C}{>{\centering\arraybackslash}p{3.5em}}
\begin{tabular}{CCCCCC}
\includegraphics[width=0.09\textwidth]{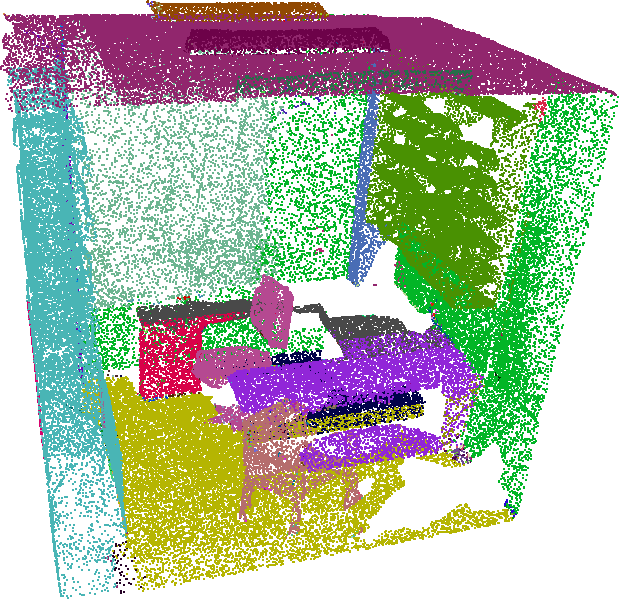}
&\includegraphics[width=0.09\textwidth]{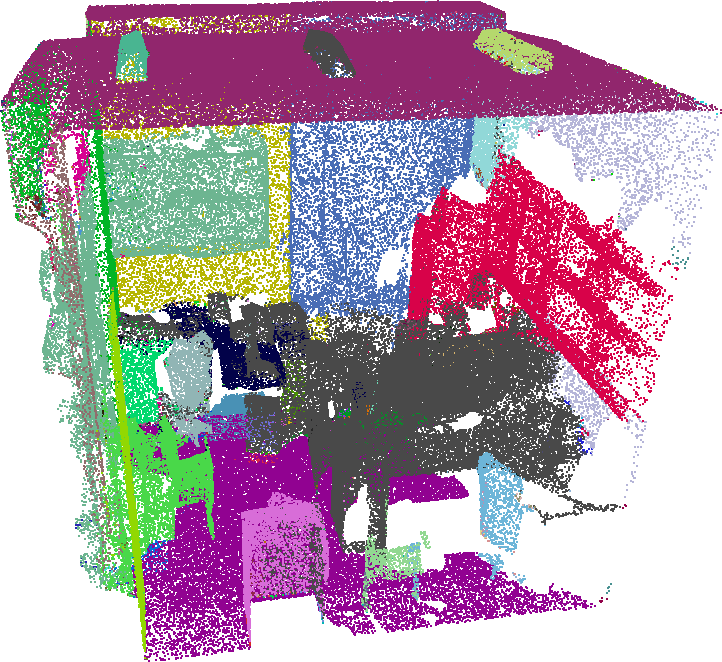}
&\includegraphics[width=0.09\textwidth]{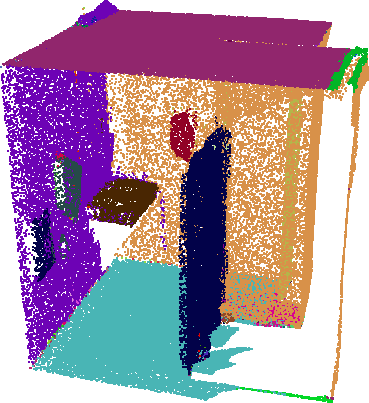}
&\includegraphics[width=0.09\textwidth]{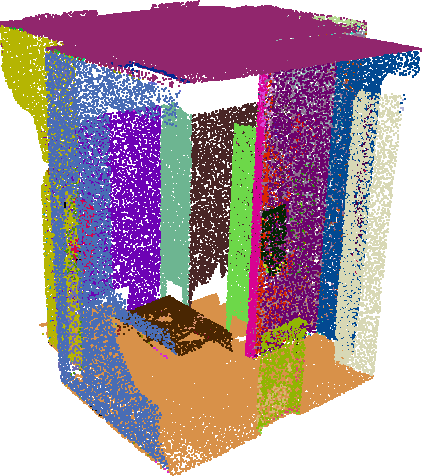}
&\includegraphics[width=0.09\textwidth]{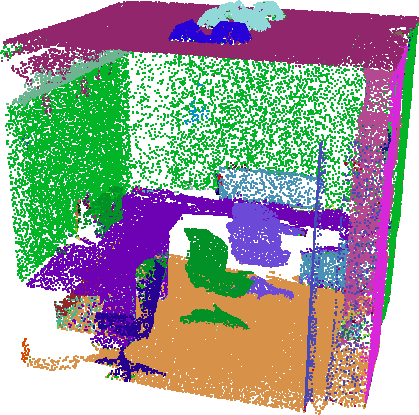}
\\
\includegraphics[width=0.09\textwidth]{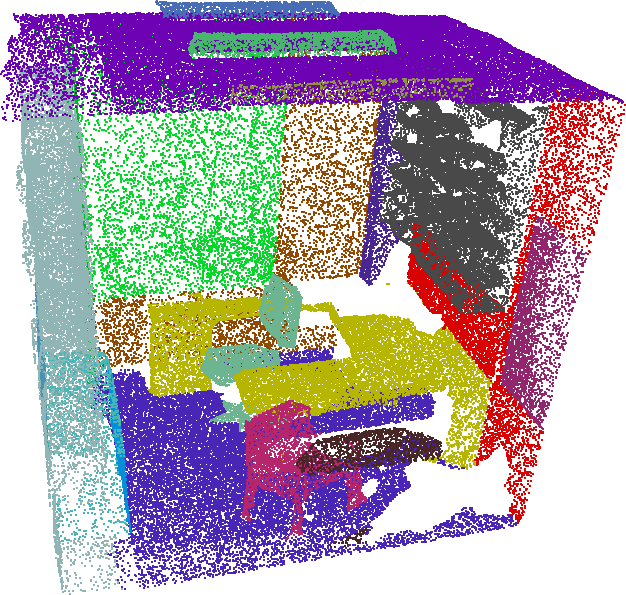}
&\includegraphics[width=0.09\textwidth]{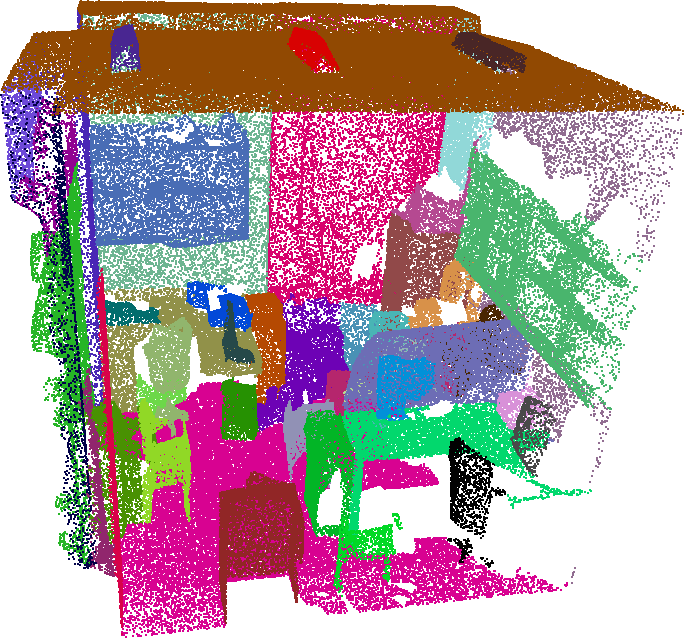}
&\includegraphics[width=0.09\textwidth]{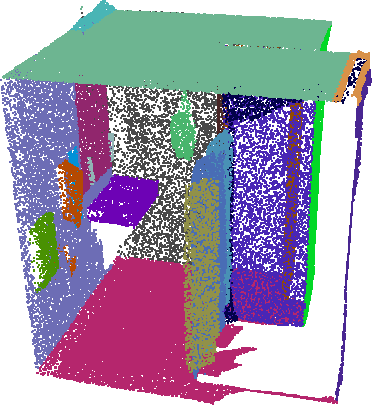}
&\includegraphics[width=0.09\textwidth]{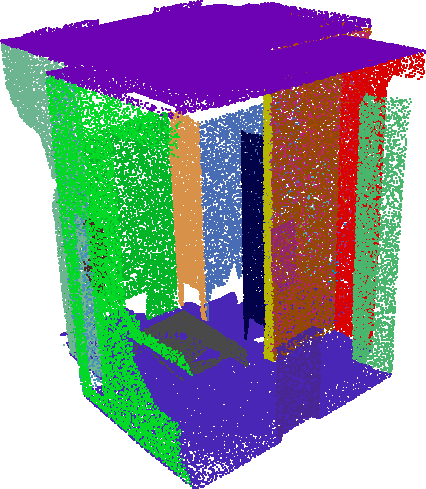}
&\includegraphics[width=0.09\textwidth]{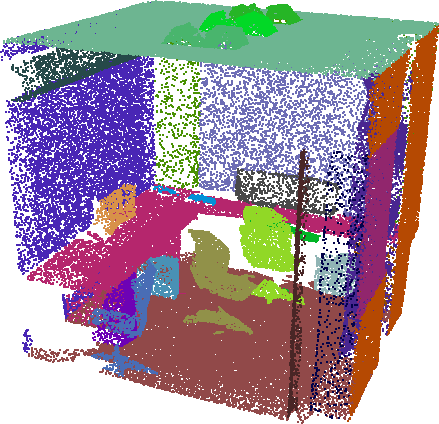}
\\
\includegraphics[width=0.09\textwidth]{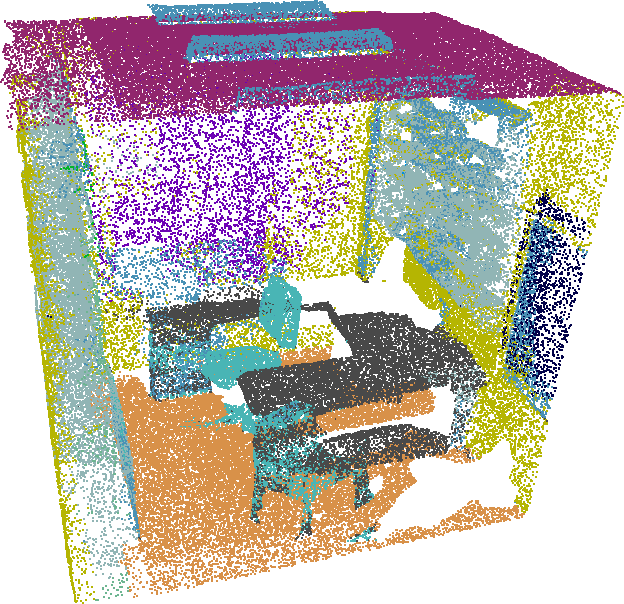}
&\includegraphics[width=0.09\textwidth]{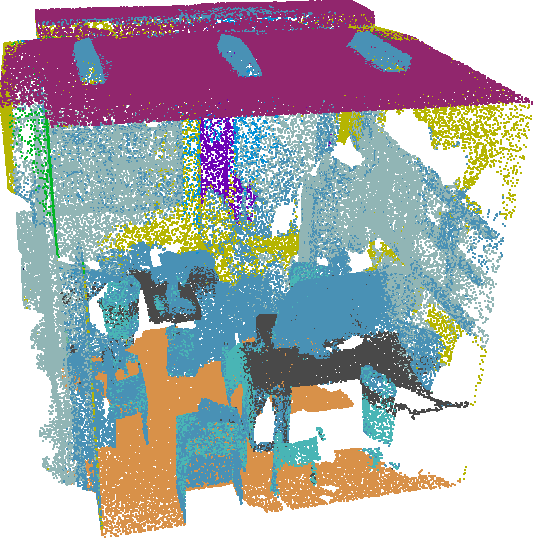}
&\includegraphics[width=0.09\textwidth]{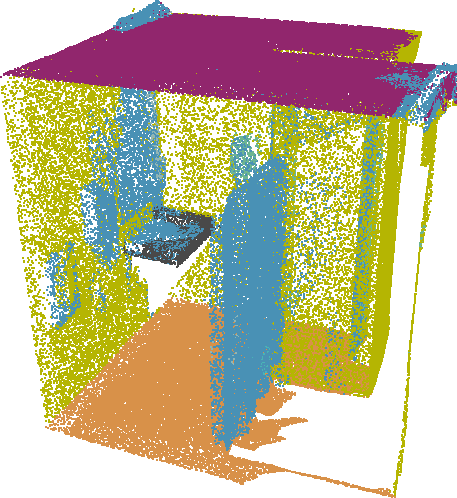}
&\includegraphics[width=0.09\textwidth]{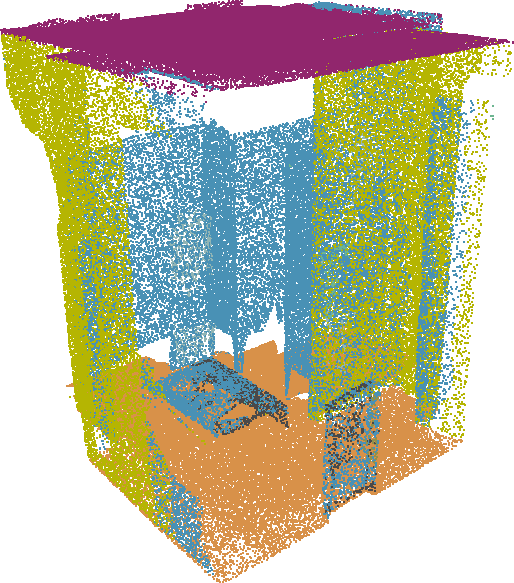}
&\includegraphics[width=0.09\textwidth]{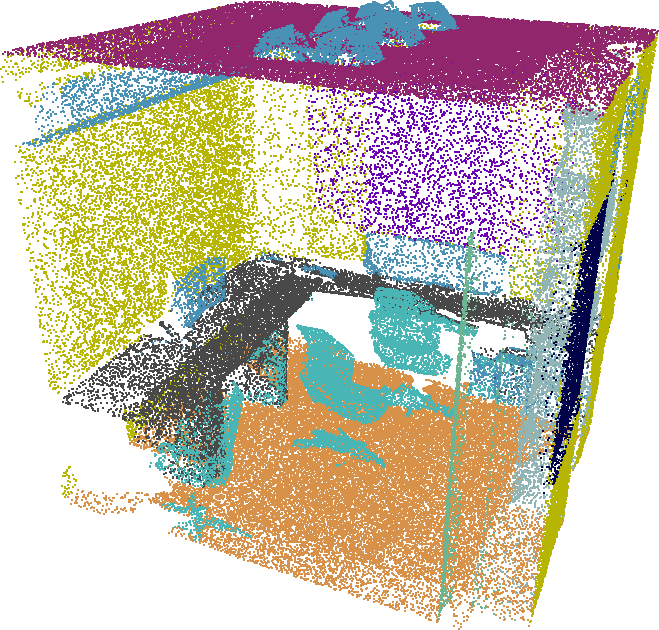}
\\
\includegraphics[width=0.09\textwidth]{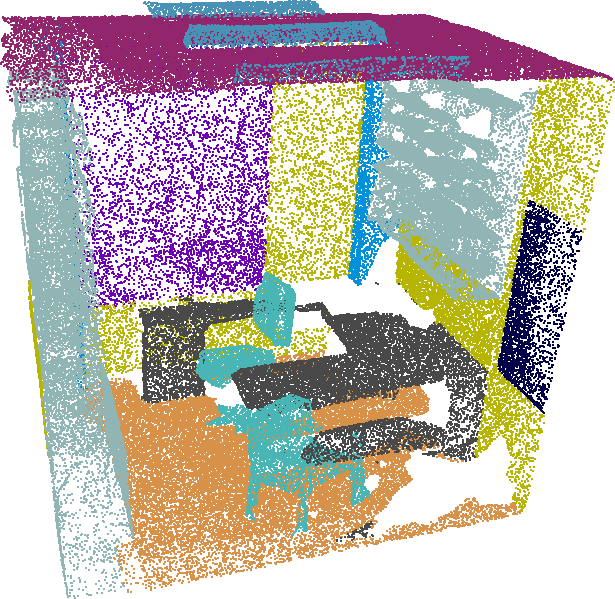}
&\includegraphics[width=0.09\textwidth]{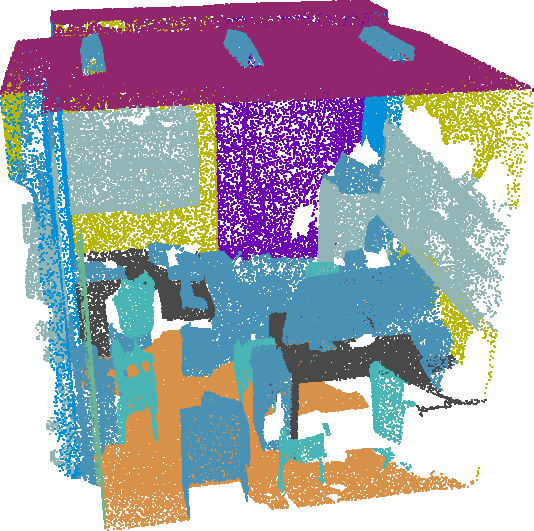}
&\includegraphics[width=0.09\textwidth]{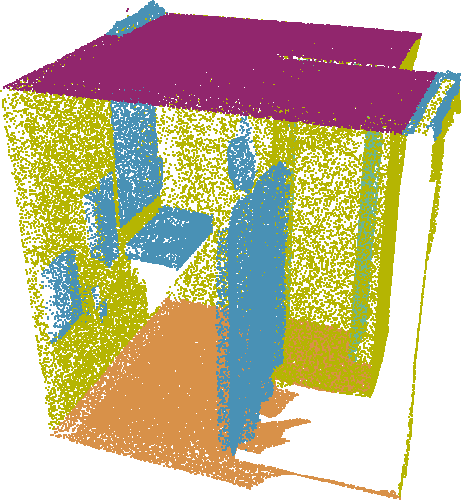}
&\includegraphics[width=0.09\textwidth]{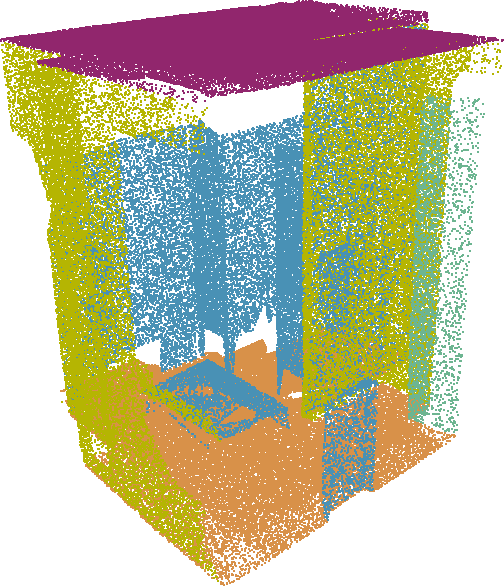}
&\includegraphics[width=0.09\textwidth]{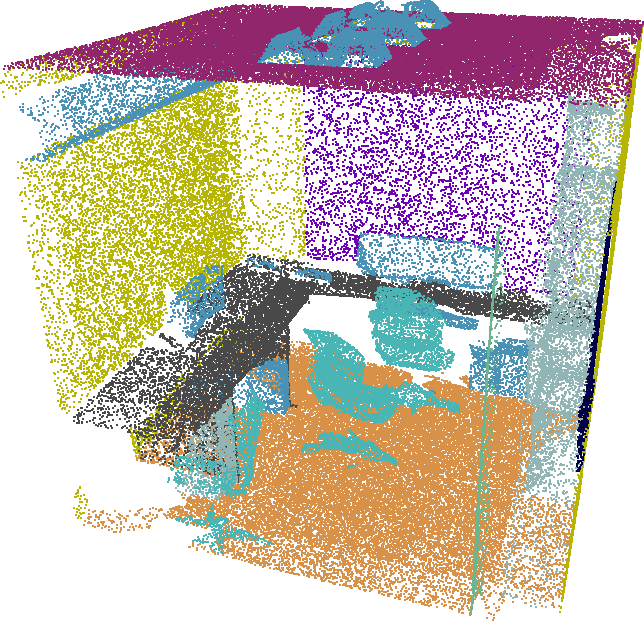}
\end{tabular}
\caption{SGPN instance segmentation results on S3DIS. The first row is the prediction results. The second row is groud truths. Different colors represent different instances. The third row is the predicted semantic segmentation results. The fourth row is the ground truths for semantic segmentation.}
\label{fig:S3DISins}
\end{figure}

\begin{figure}[tb]
\centering
\includegraphics[width=0.45\textwidth]{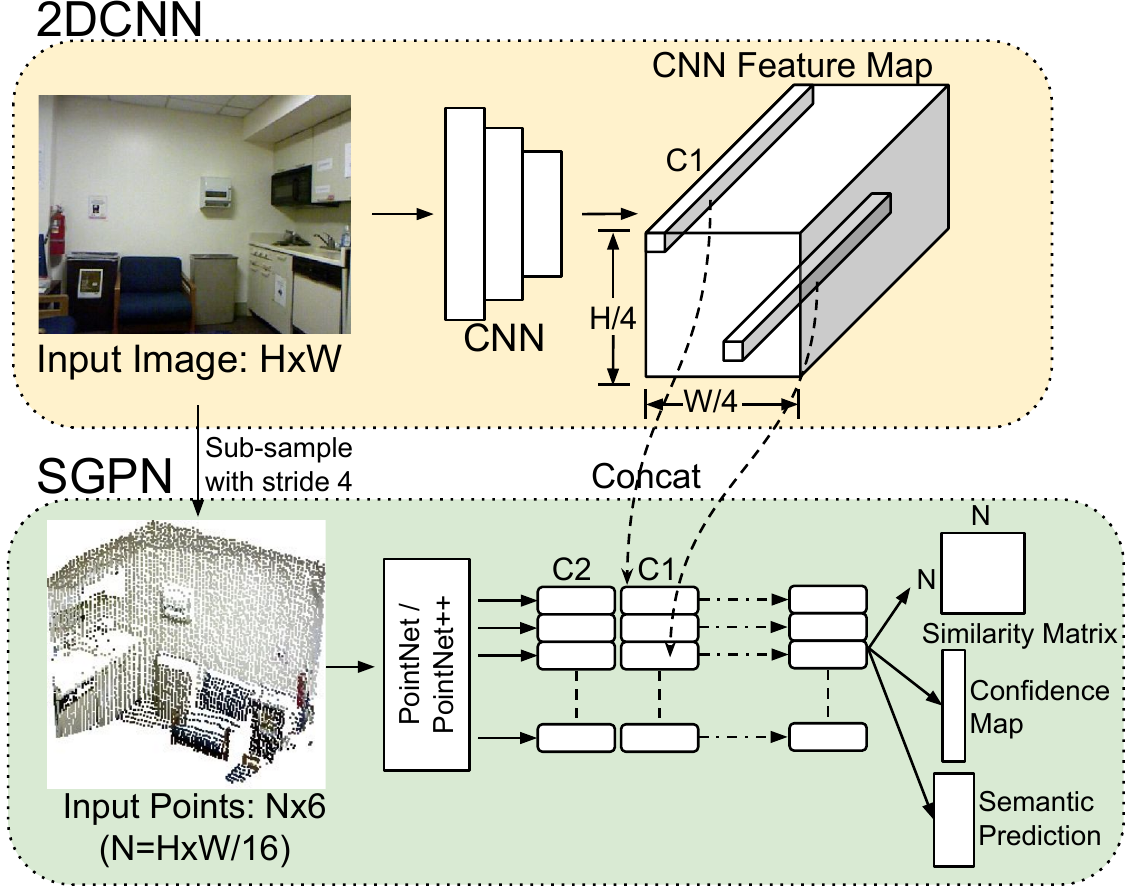}
\caption{Incorporating CNN features in SGPN.}
\label{fig:nyucnn}
\end{figure}

\begin{figure}[tb]
\centering
\newcolumntype{C}{>{\centering\arraybackslash}p{4.15em}}
\begin{tabular}{CCCCCC}
\includegraphics[width=0.10\textwidth]{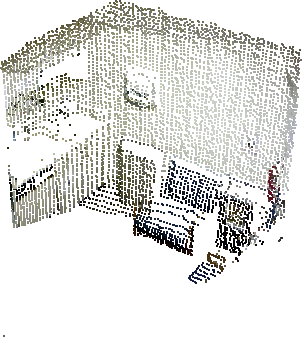}
&\includegraphics[width=0.1\textwidth]{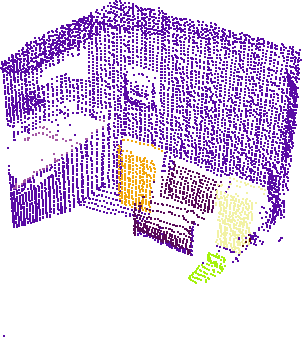}
&\includegraphics[width=0.1\textwidth]{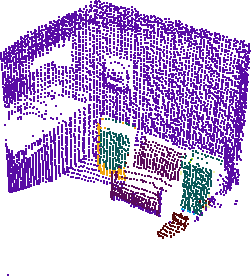}
&\includegraphics[width=0.1\textwidth]{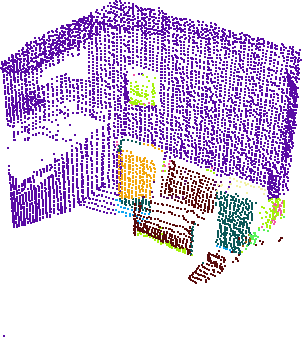}
\\
\includegraphics[width=0.1\textwidth]{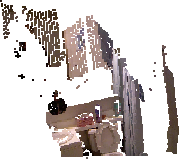}
&\includegraphics[width=0.1\textwidth]{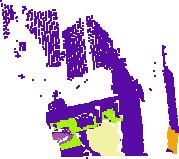}
&\includegraphics[width=0.1\textwidth]{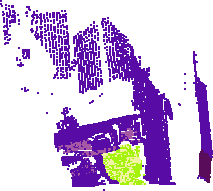}
&\includegraphics[width=0.1\textwidth]{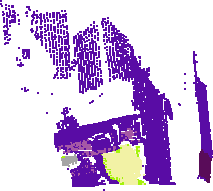}
\\
\includegraphics[width=0.1\textwidth]{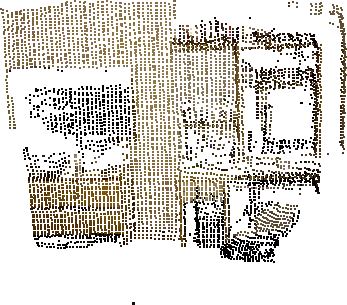}
&\includegraphics[width=0.1\textwidth]{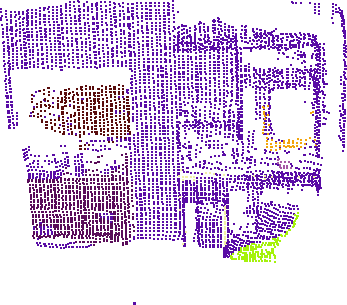}
&\includegraphics[width=0.1\textwidth]{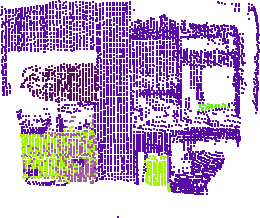}
&\includegraphics[width=0.1\textwidth]{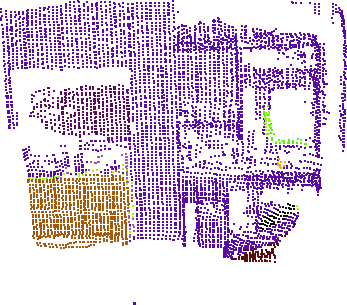}
\\
(a)&(b)&(c)&(d)
\end{tabular}
\caption{SGPN instance segmentation results on NYUV2. (a) Input point clouds. (b) Ground truths for instance segmentation. (c) Instance segmentation results with SGPN. (d) Instance segmentation results with SGPN-CNN.}
\label{fig:nyuins}
\end{figure}

\begin{table*}
\begin{center}
\setlength{\tabcolsep}{0.21em}
\newcolumntype{C}{>{\centering\arraybackslash}p{1.85em}}
\begin{tabular}{c|c|CCCCCCCCCCCCCCCCCCC}
\Xhline{3\arrayrulewidth}
& \small Mean& \multicolumn{19}{c}{\includegraphics[width=0.86\textwidth]{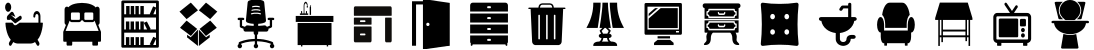}}\\
\hline
\small Seg-Cluster& 23.2&31.0&	\bf{70.1}&	27.1&	\bf{1.3}&	25.8&	20.3&	13.9&	11.1&	24.3&	4.4&	16.3&	3.6&	32.0&	25.5&	36.3&	50.9&	12.9&	2.2&23.5\\
\small SGPN  & 26.5& 55.9	&53.3&	27.8&	0.0&	27.4&	\bf{59.6}&\bf{28.9}&	6.1&	33.9&	2.0&	19.7&	2.0&	29.4&	30.7&	39.1&	43.6&	17.6&	1.2 &25.9\\
\small SGPN-CNN  &\bf{30.5}& \bf{56.4}&	55.4&	\bf{35.2}&	0.0	&\bf{42.6}	&50.6&	23.1&	\bf{21.1}&	\bf{31.8}&\bf{	7.5}&	\bf{22.7}&	\bf{6.4}	&\bf{39.9}&	\bf{33.5}&	\bf{42.4}&	\bf{54.8}&	\bf{21.3}&	\bf{3.8}&\bf{32.1}\\
  \Xhline{3\arrayrulewidth}
\end{tabular}
\end{center}
\vspace{-3mm}
\caption{Results on instance segmentation in NYUV2. The metric is AP with IoU $0.5$.}
\vspace{-4mm}
\label{table:nyuins}
\end{table*}

\begin{table}
\begin{center}
\setlength{\tabcolsep}{0.1em}
\newcolumntype{C}{>{\centering\arraybackslash}p{2.5em}}
\begin{tabular}{c|C|C|C|C|C}
\Xhline{3\arrayrulewidth}
& \small Mean
&\includegraphics[width=0.038\textwidth]{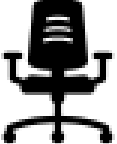}
&\includegraphics[width=0.04\textwidth]{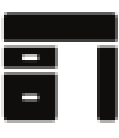}
&\includegraphics[width=0.04\textwidth]{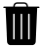}
&\includegraphics[width=0.04\textwidth]{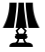}\\
\hline
Deep Sliding Shapes~\cite{DeepSlidingShapes}&37.55 & \bf{58.2} & \bf{36.1} & 27.2 & 28.7\\
Deng and Latecki~\cite{zhuo17amodal3det} &35.55 & 46.4 & 33.1 & 33.3 & 29.4 \\
SGPN & 36.25&44.4&30.4 &46.1 &24.4 \\
SGPN-CNN & \bf{41.30}&50.8&34.8 &\bf{49.4}&\bf{30.2} \\
  \Xhline{3\arrayrulewidth}
\end{tabular}
\end{center}
\vspace{-3mm}
\caption{Comparison results on 3D detection (AP with IoU 0.5) in NYUV2. Please note we use point groups as inference while ~\cite{DeepSlidingShapes,zhuo17amodal3det} use large bounding box with invisible regions as ground truth. Our prediction is the tight bounding box on points which makes the IoU much smaller than ~\cite{DeepSlidingShapes,zhuo17amodal3det}.}
\vspace{-2mm}
\label{table:nyuv2det}
\end{table}

\subsection{NYUV2 Object Detection and Instance Segmentation Evaluation}  \label{nyu}
We  evaluate the effectiveness of our approach on partial 3D scans on the NYUV2 dataset. In this dataset, 3D point clouds are lifted from a single RGBD image. An image of size $H\times W$ can produce $H\times W$ points. We subsample this point cloud by resizing the image to $\frac{H}{4} \times \frac{W}{4}$ and get the corresponding points using a nearest neighbor search. Both our training and testing experiments are conducted on such a  point cloud. PointNet++ is used as our baseline.

In ~\cite{qi20173d}, 2D CNN features are combined 3D point cloud for RGBD semantic segmentation. By leveraging the \textit{flexibility} of SGPN, we also seamlessly integrate 2D CNN features from RGB images to boost performance. A 2D CNN consumes an RGBD map and extracts feature maps $F_2$  with size $\frac{H}{4} \times \frac{W}{4} \times N_{F2}$. Since there are $\frac{H}{4} \times \frac{W}{4}$ sub-sampled points for every image, a feature vector of size $N_{f2}$ can be extracted from $F_2$ at each pixel location. Every feature vector is concatenated to $F$ (a $N_p\times N_F$ feature matrix produced by PointNet/PointNet++ as mentioned in Section~\ref{sec:sgpn}) for each corresponding point, yielding a feature map of size $N_P\times (N_F+N_{F2})$, which we then feed to our output branches. Figure~\ref{fig:nyucnn} illustrates this procedure; we call this pipeline SGPN-CNN. In our experiments, we use a pre-trained AlexNet model ~\cite{krizhevsky2012imagenet} (with the first layer stride $1$) and extract $F_2$ from the \texttt{conv5} layer. We use $H\times W = 316\times415$ and $N_p = 8137$. The 2D CNN and SGPN are trained jointly.

\begin{table*}[h!]
\begin{center}
\setlength{\tabcolsep}{0.1em}
\newcolumntype{C}{>{\centering\arraybackslash}p{2.2em}}
\begin{tabular}{c|C|CCCCCCCCCCCCCCCCC}
\Xhline{3\arrayrulewidth}
& \small Mean& \thead{air-\\plane}& bag& cap& car& chair& \thead{head \\ phone}& guitar& knife& lamp& laptop& motor& mug& pistol& rocket& \thead{skate \\ board}& table\\
\hline
~\cite{pointnet2}& 84.6& \bf{80.4} & \bf{80.9}&  60.0   &     \bf{76.8} & 88.1  & \bf{83.7}& 90.2&  82.6&  76.9&  94.7 & 68.0 & 91.2 &  82.1&  59.9 & 78.2&  87.5\\
SGPN& \bf{85.8} & \bf{80.4}  & 78.6  &            \bf{78.8} & 71.5 & \bf{88.6} & 78.0&  \bf{90.9}&  \bf{83.0} &  \bf{78.8}&  \bf{95.8} &  \bf{77.8} & \bf{93.8}&  \bf{87.4}&  \bf{60.1} & \bf{92.3}&  \bf{89.4}\\

  \Xhline{3\arrayrulewidth}
\end{tabular}
\end{center}
\vspace{-3mm}
\caption{Semantic segmentation results on ShapeNet part dataset. Metric is mean IoU(\%) on points.}
\label{table:shapenetsemseg}
\end{table*}

\begin{figure}[tb]
\centering
\renewcommand{\arraystretch}{0.1}
\begin{tabular}{cccccc}
(a)&\includegraphics[width=0.06\textwidth]{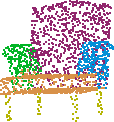}
&\includegraphics[width=0.09\textwidth]{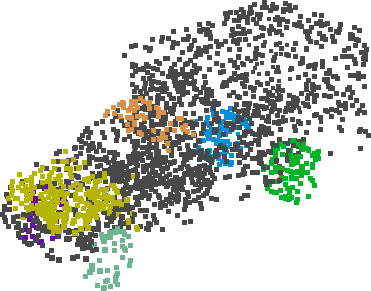}
&\includegraphics[width=0.09\textwidth]{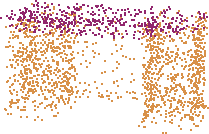}
&\includegraphics[width=0.05\textwidth]{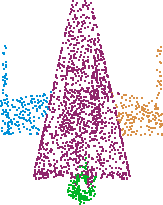}
&\includegraphics[width=0.012\textwidth]{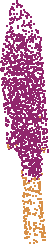}
\\
(b)&\includegraphics[width=0.06\textwidth]{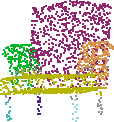}
&\includegraphics[width=0.09\textwidth]{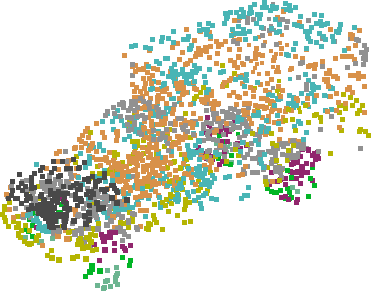}
&\includegraphics[width=0.09\textwidth]{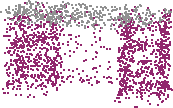}
&\includegraphics[width=0.05\textwidth]{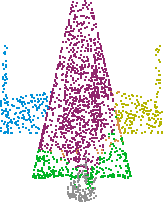}
&\includegraphics[width=0.012\textwidth]{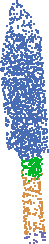}
\\
(c)&\includegraphics[width=0.06\textwidth]{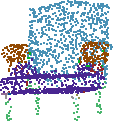}
&\includegraphics[width=0.09\textwidth]{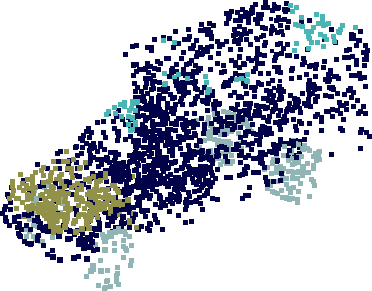}
&\includegraphics[width=0.09\textwidth]{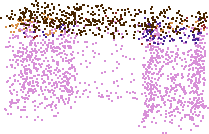}
&\includegraphics[width=0.05\textwidth]{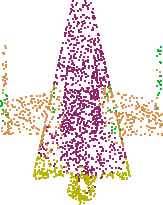}
&\includegraphics[width=0.012\textwidth]{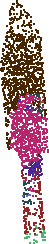}
\\
(d)&\includegraphics[width=0.06\textwidth]{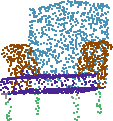}
&\includegraphics[width=0.09\textwidth]{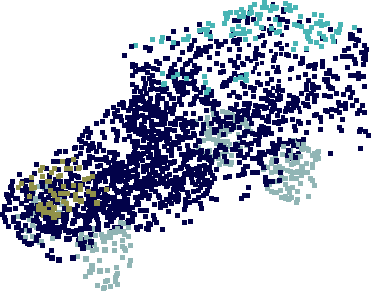}
&\includegraphics[width=0.09\textwidth]{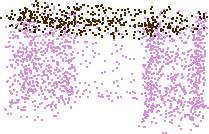}
&\includegraphics[width=0.05\textwidth]{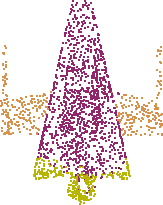}
&\includegraphics[width=0.012\textwidth]{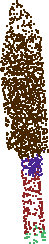}

\end{tabular}
\caption{Qualitative results on ShapeNet Part Dataset. (a) Generated ground truth for instance segmentation. (b) SGPN instance segmentation results. (c) Semantic segmentation results of PointNet++. (d) Semantic segmentation results of SGPN.}
\label{fig:shapenetpart}
\end{figure}

Evaluation is performed on 19 object categories. Figure~\ref{fig:nyuins} shows qualitative results on instance segmentation of SGPN. Table~\ref{table:nyuins} shows comparisons between Seg-Cluster and our SGPN and CNN-SGPN frameworks on instance segmentation. The evaluation metric is average precision (AP) with IoU threshold $0.5$.

The margin of improvement for SGPN compared to Seg-Cluster is not as high as it is on S3DIS, because in this dataset objects with the same semantic label are usually far apart in Euclidean space. Additionally, naive methods like Seg-Cluster benefit since it is easy to separate a single instance into parts since the points are not connected due to occlusion in partial scanning. Table~\ref{table:nyuins} also illustrates that SGPN can effectively utilize CNN features. Instead of concatenating fully-connected layer of 2D and 3D networks as in ~\cite{DeepSlidingShapes}, we combine 2D and 3D features by considering their geometric relationships.

We further calculate bounding boxes with instance segmentation results. Table \ref{table:nyuv2det}  compares our work with the state-of-the-art works~\cite{DeepSlidingShapes,zhuo17amodal3det} on NYUV2 3D object detection. Following the evaluation metric in ~\cite{slidingshapes}, AP is calculated with IoU threshold $0.25$ on 3D bounding boxes. The NYUV2 dataset provides ground truth 3D bounding boxes that encapsulate the whole object including the part that is invisible in the depth image. Both ~\cite{DeepSlidingShapes} and ~\cite{zhuo17amodal3det} use these large ground truth bounding boxes for inference. In our method, we infer point groupings, which lack information of the invisible part of the object. Our output is the derived tight bounding box around the grouped points in the partial scan, which makes our IoUs much smaller than ~\cite{DeepSlidingShapes,zhuo17amodal3det}. However, we can still see the effectiveness of SGPN on the task of 3D object detection on partial scans as our method achieves better performance on small objects.
\vspace{-3mm}
\paragraph{Computation Speed} To benchmark the testing time with ~\cite{DeepSlidingShapes,zhuo17amodal3det} and make fair comparison, we run our framework on an Nvidia K40 GPU. SGPN takes $170$ms and around $400$M GPU memory per sample. CNN-SGPN takes $300$ms and $1.4$G GPU memory per sample. \texttt{GroupMerging} takes $180$ms on an Intel i7 CPU. However, the detection net in ~\cite{zhuo17amodal3det} takes $739$ms on an Nvidia Titan X GPU. In ~\cite{DeepSlidingShapes}, RPN takes 5.62s and ORN takes 13.93s per image on an Nvidia K40 GPU. Our model improves the efficiency and reduces GPU memory usage by a large margin.

\subsection{ShapeNet Part Instance Segmentation}
\label{sec:experimentshapenet}
Following the settings in ~\cite{pointnet2}, point clouds are generated by uniformly sampling shapes from Shapenet~\cite{shapenet}. In our experiments we sample each shape into $2048$ points. The XYZ of points are fed into network as input with size $2048\times 3$. To generate ground truth labels for part instance segmentation from semantic segmentation results, we perform DBSCAN clustering on each part category of an object to group points into instances. This experiment is conducted as a toy example to demonstrate the effectiveness of our approach on instance segmentation for pointclouds.

We use Pointnet++ as our baseline. Figure~\ref{fig:shapenetpart}(b) illustrates the instance segmentation results. For instance results, we again use different colors to represent different instances, and point colors of the same group are not necessarily the same as the ground truth.  Since the generated ground truths are not ``real" ground truths, only qualitative results are provided. SGPN achieves good results even under challenging conditions. As we can see from the Figure~\ref{fig:shapenetpart}, SGPN is able to group the chair legs into four instances even though even in the ground truth DBSCAN can not separate the chair legs apart.

The similarity matrix can also help the semantic segmentation branch training. We compare SGPN to PointNet++ (i.e. our framework with solely a semantic segmentation branch) on semantic segmentation in Table~\ref{table:shapenetsemseg}. The inputs of both networks are point clouds of size $2048\time 3$. Evaluation metric is mIoU on points of each shape category. Our model performs better than PointNet++ due to the similarity matrix. Qualitative results are shown in Figure~\ref{fig:shapenetpart}. Some false segmentation prediction is refined with the help of SGPN.



%% file: supp.tex
\section{Network Architecture}
In our experiments, we use both PointNet and PointNet++ as our baseline architectures. For the S3DIS dataset, we use PointNet as our baseline for fair comparison with the 3D object detection system described in the PointNet paper~\cite{pointnet}. The network architecture is the same as the semantic segmentation network as stated in PointNet except for the last two layers. Our $F$ is the last $1\times 1$ $conv$ layer with BatchNorm and ReLU in PointNet with $256$ output channels. $F_{SIM}, F_{CF}, F_{SEM}$ are $1\times 1$ $conv$ layers with output channels $(128, 128, 128)$, respectively. 

For the NYUV2 dataset, we use PointNet++ as our baseline. We use the same notations as PointNet++ to describe our architecture: 

$SA(K,r,[l_1, ..., l_d])$ is a set abstraction ($SA$) level with $K$ local regions of ball radius $r$ using a PointNet
architecture of $d$ $1\times 1$ $conv$ layers with output channels $l_i (i = 1, ..., d)$. $FP(l_1, ..., l_d)$ is a feature
propagation ($FP$) level with $d$ $1\times 1$ $conv$ layers.
Our network architecture is: 
\begin{flalign*}
&SA(1024, 0.1, [32, 32, 64]), \\
&SA(256, 0.2, [64, 64, 128]), \\
&SA(128, 0.4, [128, 128, 256]), \\
&SA(64, 0.8, [256, 256, 256]), \\
&SA(16, 1.2, [256, 256, 512]), \\
&FP(512, 256), \\
&FP(256, 256), \\
&FP(256, 256), \\
&FP(256, 128), \\
&FP(128, 128, 128, 128).
\end{flalign*}
$F_{SIM}, F_{CF}, F_{SEM}$ are $1\times 1$ $conv$ layers with output channels $(128, 128, 128)$ respectively.

For our experiments on the ShapeNet part dataset, PointNet++ is used as our baseline. We use the same network architecture as in the PointNet++ paper~\cite{pointnet2}. $F_{SIM}, F_{CF}, F_{SEM}$ are $1\times 1$ $conv$ layers with output channels $(64, 64, 64)$, respectively.

%
%
\section{Experiment Settings}
\subsection{S3DIS Dataset}
\begin{figure}[tb]
	\centering
	\includegraphics[width=0.35\textwidth]{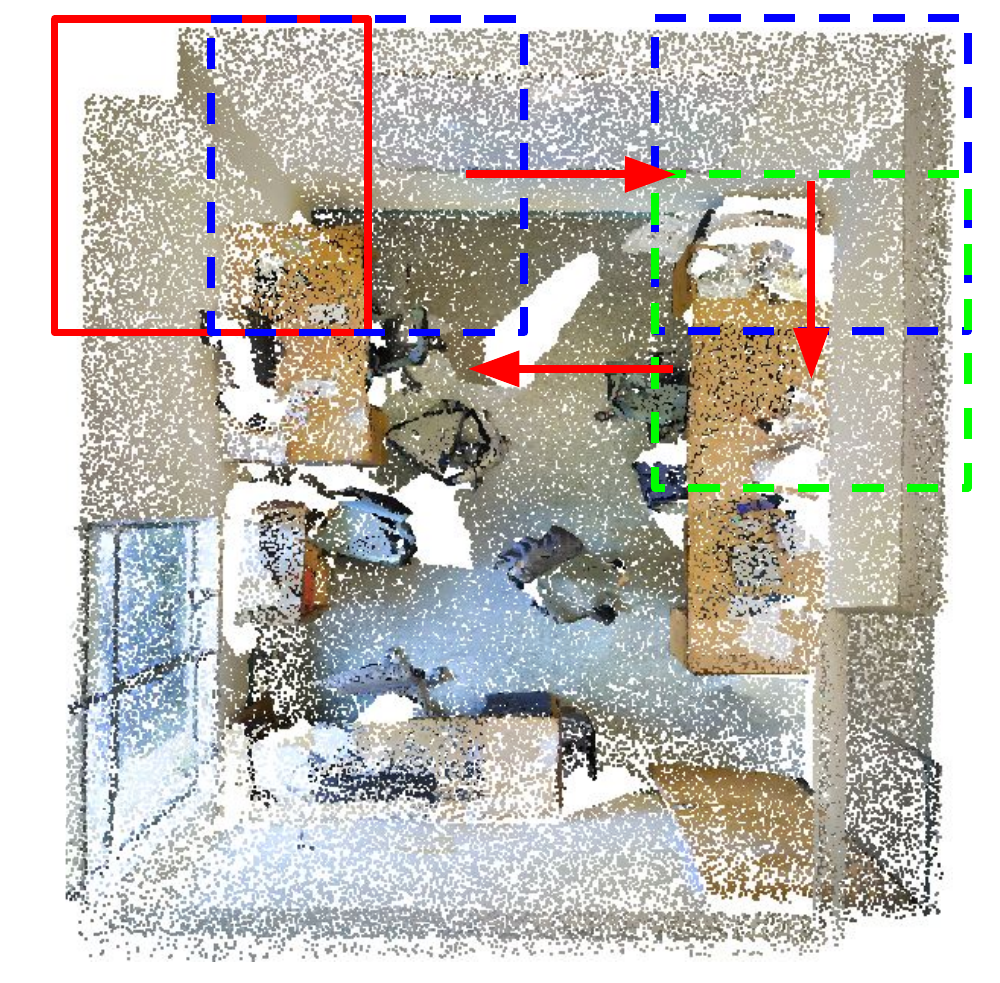}
	\caption{Dividing scene into blocks with overlap (top view).}
	\label{fig:block}
\end{figure}

\paragraph{Block Merging}
We divide each scene into $1m \times 1m$ blocks with overlapping sliding windows in a snake pattern of stride $0.5m$ as is shown in Figure~\ref{fig:block}. The entire scene is also divided into a $400\times400\times400$ grid $V$. $V_k$ is used to indicate the instance label of cell $k$ where $k \in [0,400\times400\times400) $. Given $V$ and point instance labels for each block $PL$ where $PL_{ij}$ represents the instance label of $j$th point in block $i$, a \textit{BlockMerging} algorithm (refer to Algorithm~\ref{alg:block}) is derived to merge object instances from different blocks. 

\begin{table*}[ht]
	\begin{center}
		\setlength{\tabcolsep}{0.21em}
		\newcolumntype{C}{>{\centering\arraybackslash}p{2em}}
		\newcolumntype{E}{>{\centering\arraybackslash}p{2.5em}}
		\newcolumntype{D}{>{\centering\arraybackslash}p{1.8em}}
		\begin{tabular}{c|C|DDDDDDDDDDCCCDDECDD}
			\Xhline{3\arrayrulewidth}
			& \small Mean & \thead{wall} & \thead{floor} & \thead{chair} & \thead{table} & \thead{desk} & \thead{bed} & \thead{book \\ shelf} & \thead{sofa} & \thead{sink} &  \thead{bath\\tub} & \thead{toilet} & \thead{cur-\\tain} & \thead{coun-\\ter} & \thead{door} & \thead{win-\\dow} & \thead{shower\\curtain} & \thead{fridge} &\thead{pic-\\ture} &\thead{cabi-\\net}
			\\
			\hline
			Seg-Cluster&32.5&	34.2&\bf{87.6}&	61.4&40.4&20.7&\bf{43.1}&19.2&33.8&	33.5&	44.4&	\bf{60.0}&	\bf{39.8}&	24.5&	0&	0&	\bf{19.4}&22.0&	0&	33.5	\\
			SGPN& \textbf{35.1}&\bf{55.5}&86.7&	\bf{64.4}&\bf{41.1}&	\bf{40.7}&42.7&	\bf{36.1}&\bf{39.6}&	\bf{38.6}&	\bf{45.1}&	50.3&	15.9&	\bf{27.0}&0&	0&	19.3&	\bf{28.6}&	0&	\bf{35.1}	
			\\
			\Xhline{3\arrayrulewidth}
		\end{tabular}
	\end{center}
	\caption{Instance segmentation results on ScanNet(v1). The metric is AP (\%) with IoU threshold $0.25$. We observe 0 percent AP on items that appear on the wall (door, window, picture) as they contain very little depth information and are almost all incorrectly semantically labeled as the wall. Future works can explore addressing this problem.}
	\label{table:scannet}
\end{table*}

\begin{algorithm}
	\caption{BlockMeriging}
	\label{alg:block}
	\SetKwInOut{Input}{Input}
	\SetKwInOut{Output}{Output}
	
	\Input{$V$, $PL$}
	\Output{Point instance labels for the whole scene $L$}
	Initialize $V$ with all elements $-1$\;
	$GroupCount \leftarrow 0$\;
	\For{every block $i$}
	{
		\eIf{ $i$ is the $1$st block}{
			\For{every point $P_j$ in block $i$}
			{
				Define $k$ where $P_j$ is located in the $k$th cell of $V$\;
				$V_k$ $\leftarrow PL_{1j}$\;
			}
		}
		{
			\For{every instance $I_j$ in block $i$}
			{
				Define $V_{I_j}$  points in  $I_j$ are located in cells $V_{I_j}$\;
				$V_t \leftarrow$ the cells in $V_{I_j}$ that do not have value  $-1$\;
				\eIf{the frequency of the mode in $V_t<30$}{
					$V_{I_j} \leftarrow GroupCount$\;
					$GroupCount \leftarrow GroupCount+1$\;
				}
				{
					$V_{I_j} \leftarrow$ the mode of $V_t$\;
				}	    		
			}
		}
	}
	\For{every point $P_j$ in the whole scene}{
		Define $k$ where $P_j$ is located in the $k$th cell of $V$\;
		$L_j$ $\leftarrow V_k$\;
	}
\end{algorithm}

\begin{figure*}[tb]
	\centering
	\begin{tabular}{cc|cc}
		\includegraphics[width=0.2\textwidth,height=0.13\textheight]{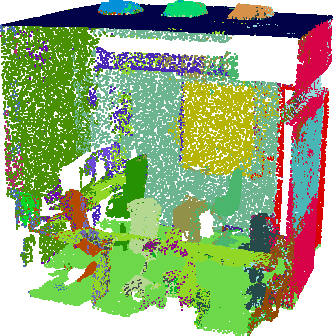}&
		\includegraphics[width=0.2\textwidth,height=0.13\textheight]{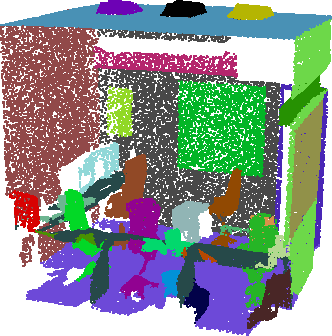}&
		\includegraphics[width=0.2\textwidth,height=0.13\textheight]{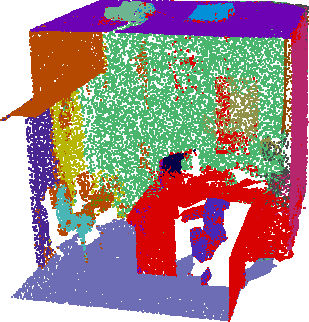}&
		\includegraphics[width=0.2\textwidth,height=0.13\textheight]{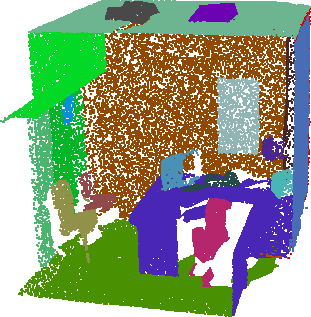}
		\\
		\includegraphics[width=0.2\textwidth,height=0.13\textheight]{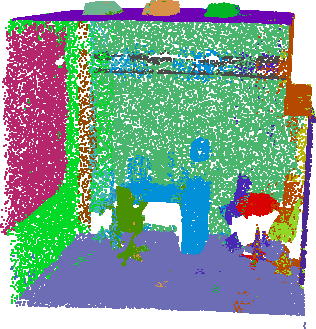}&
		\includegraphics[width=0.2\textwidth,height=0.13\textheight]{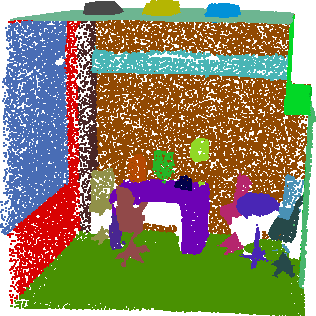}&
		\includegraphics[width=0.2\textwidth,height=0.13\textheight]{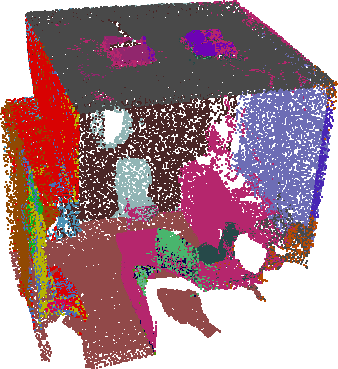}&
		\includegraphics[width=0.2\textwidth,height=0.13\textheight]{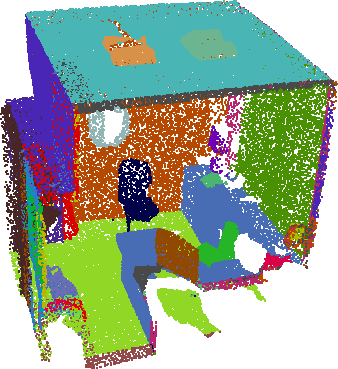}
		\\
		\includegraphics[width=0.2\textwidth,height=0.13\textheight]{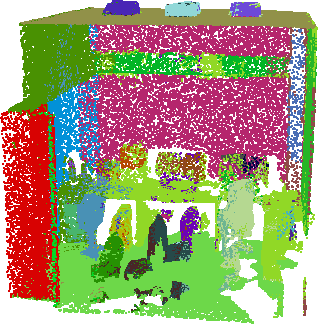}&
		\includegraphics[width=0.2\textwidth,height=0.13\textheight]{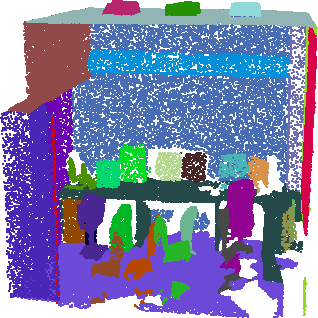}&
		\includegraphics[width=0.2\textwidth,height=0.13\textheight]{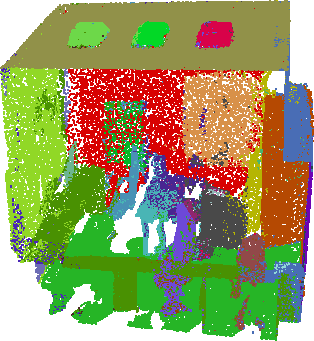}&
		\includegraphics[width=0.2\textwidth,height=0.13\textheight]{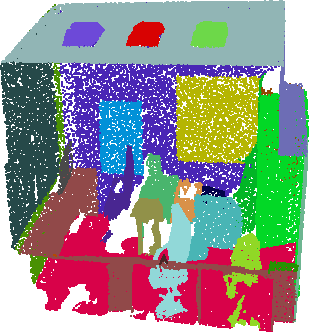}
		\\
		\includegraphics[width=0.2\textwidth,height=0.13\textheight]{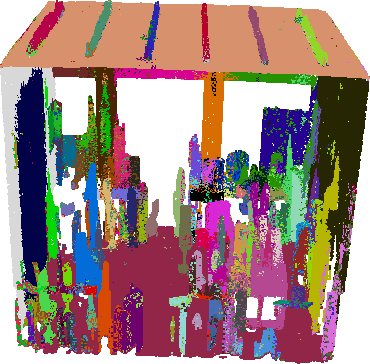}&
		\includegraphics[width=0.2\textwidth,height=0.13\textheight]{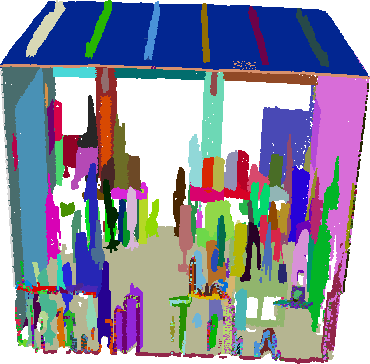}&
		\includegraphics[width=0.2\textwidth,height=0.13\textheight]{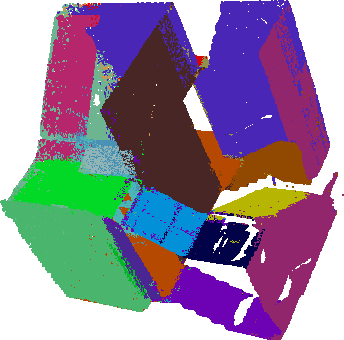}&
		\includegraphics[width=0.2\textwidth,height=0.13\textheight]{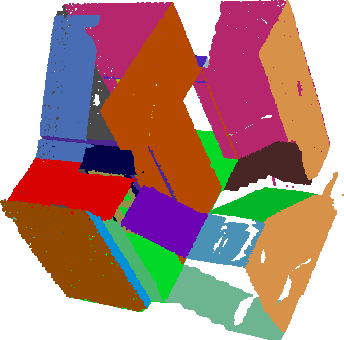}
		\\
		\includegraphics[width=0.2\textwidth,height=0.13\textheight]{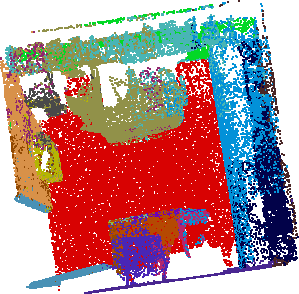}&
		\includegraphics[width=0.2\textwidth,height=0.13\textheight]{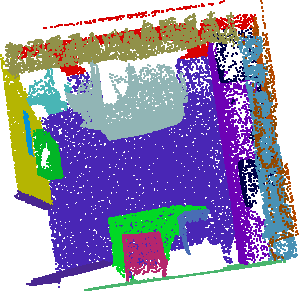}&
		\includegraphics[width=0.2\textwidth,height=0.13\textheight]{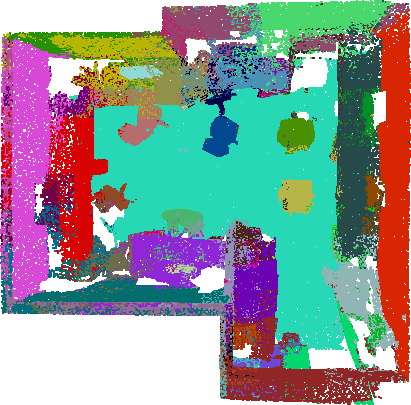}&
		\includegraphics[width=0.2\textwidth,height=0.13\textheight]{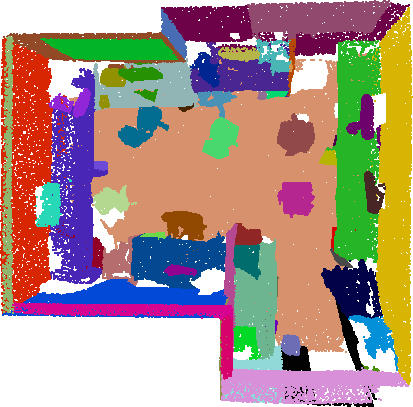}
		\\
		\includegraphics[width=0.2\textwidth,height=0.13\textheight]{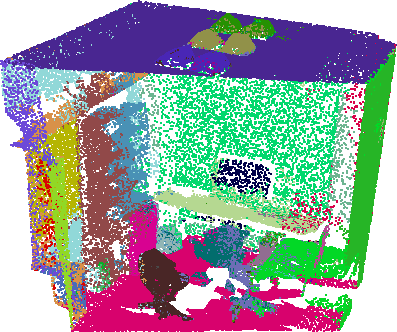}&
		\includegraphics[width=0.2\textwidth,height=0.13\textheight]{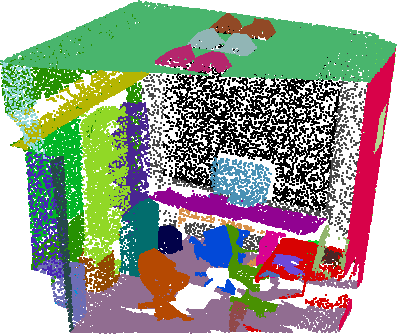}&
		\includegraphics[width=0.2\textwidth,height=0.13\textheight]{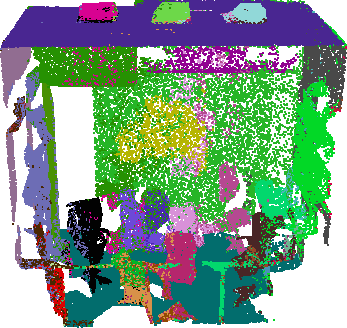}&
		\includegraphics[width=0.2\textwidth,height=0.13\textheight]{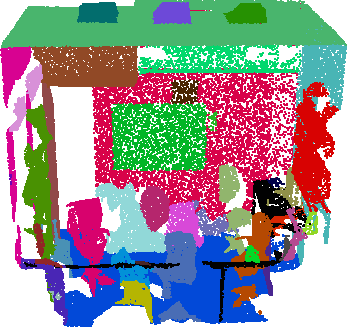}
		\\
		Prediction & Ground Truth & Prediction & Ground Truth
	\end{tabular}
	\caption{Instance segmentation results on S3DIS with SGPN. Different colors represent different instances. The colors of the same object in ground truth and prediction are not necessarily the same.}
	\label{fig:exampleblock}
\end{figure*}

In Figure~\ref{fig:exampleblock}, we show more qualitative results of instance segmentation with SGPN.

\section{More Experiments}
\subsection{ScanNet}
\begin{figure*}[tb]
	\centering
	\begin{tabular}{cc|cc}
		\includegraphics[width=0.2\textwidth,height=0.13\textheight]{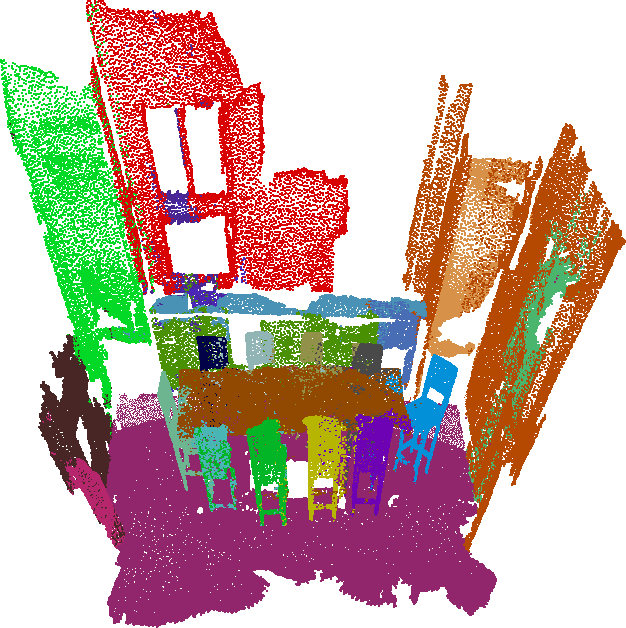}&
		\includegraphics[width=0.2\textwidth,height=0.13\textheight]{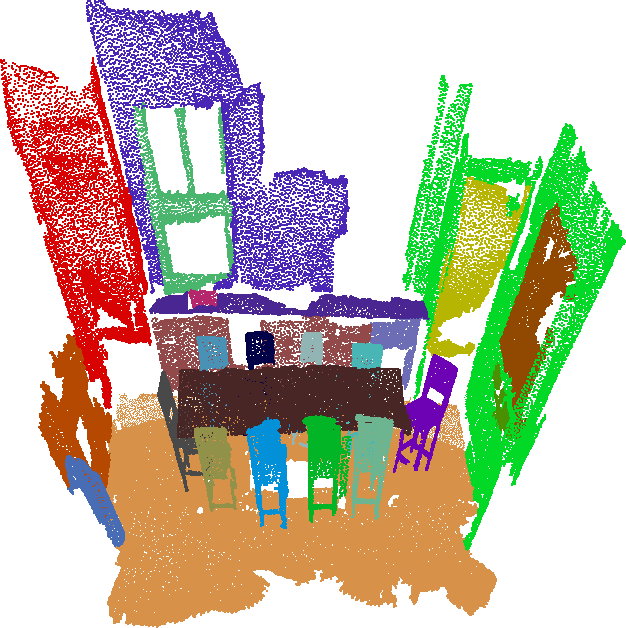}&
		\includegraphics[width=0.2\textwidth,height=0.13\textheight]{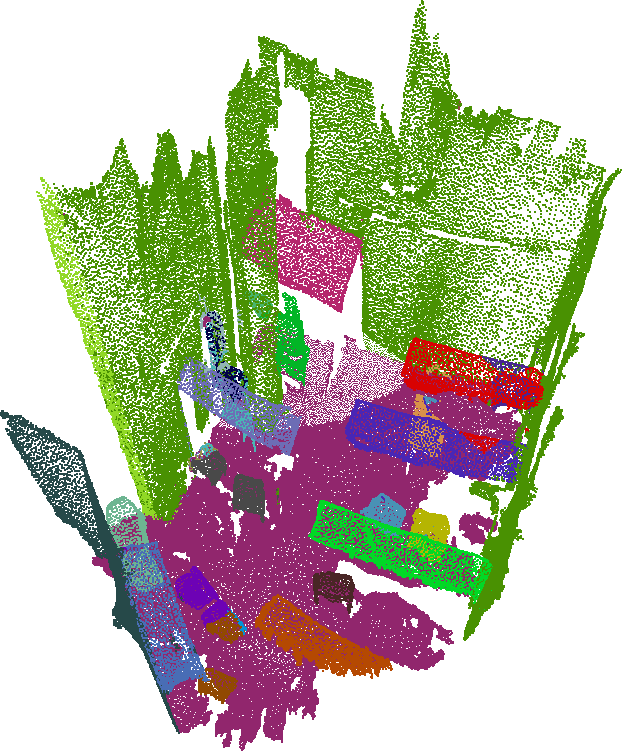}&
		\includegraphics[width=0.2\textwidth,height=0.13\textheight]{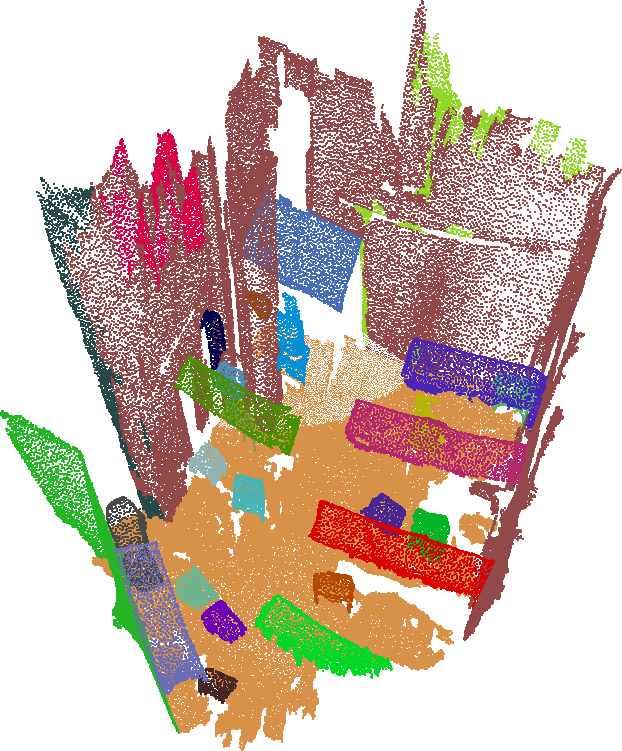}
		\\
		\includegraphics[width=0.2\textwidth,height=0.13\textheight]{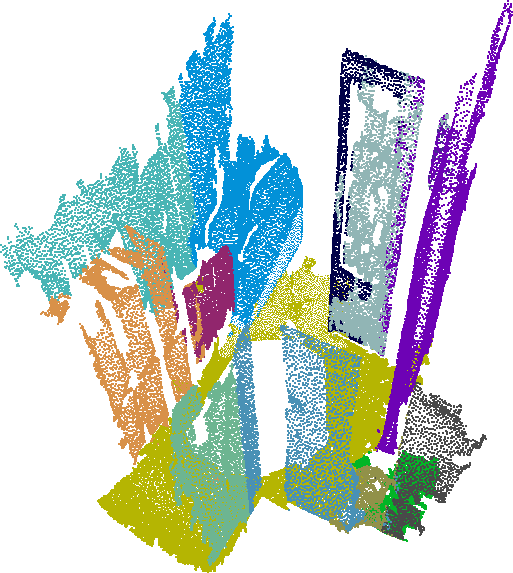}&
		\includegraphics[width=0.2\textwidth,height=0.13\textheight]{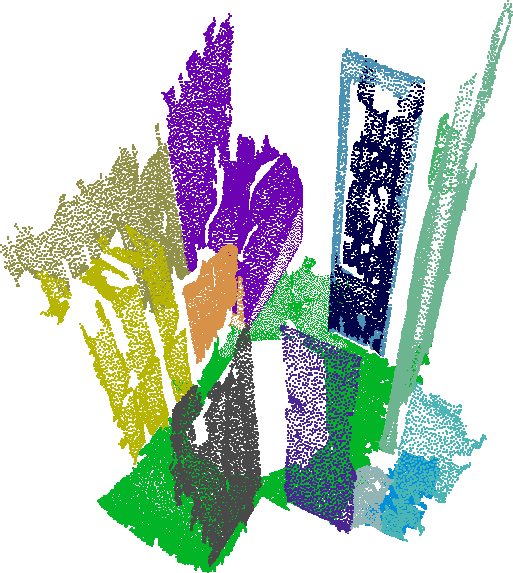}&
		\includegraphics[width=0.2\textwidth,height=0.13\textheight]{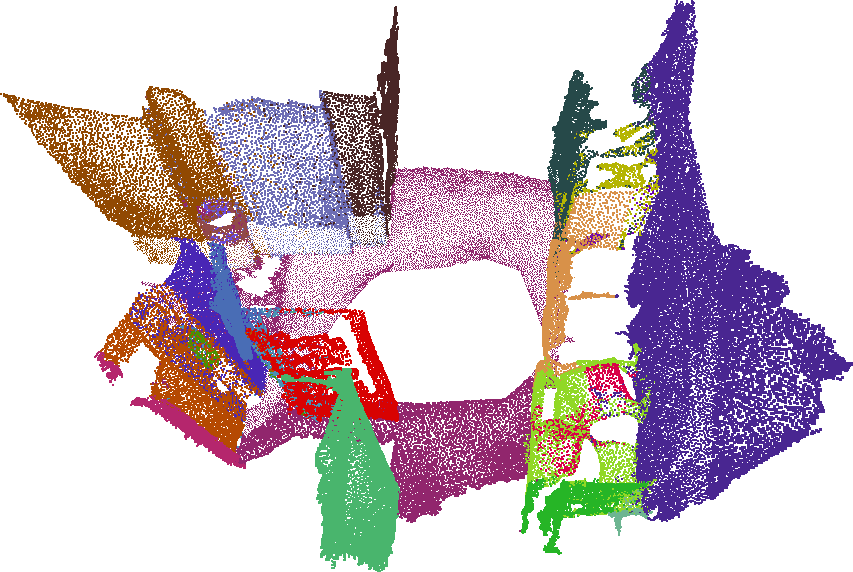}&
		\includegraphics[width=0.2\textwidth,height=0.13\textheight]{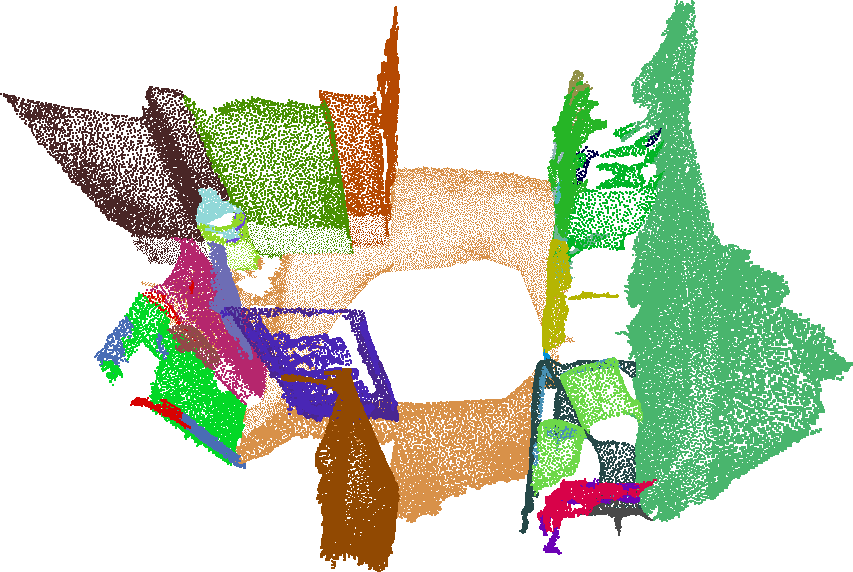}
		\\
		\includegraphics[width=0.2\textwidth,height=0.13\textheight]{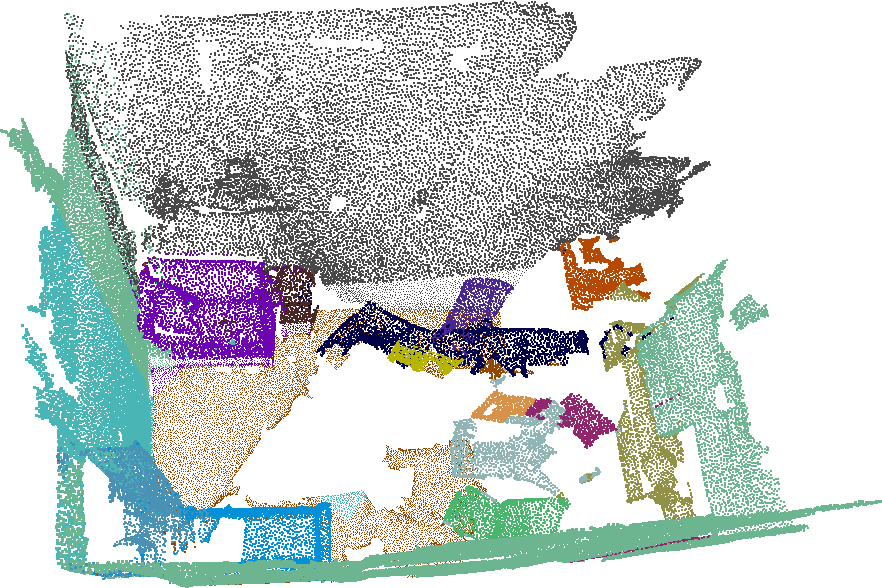}&
		\includegraphics[width=0.2\textwidth,height=0.13\textheight]{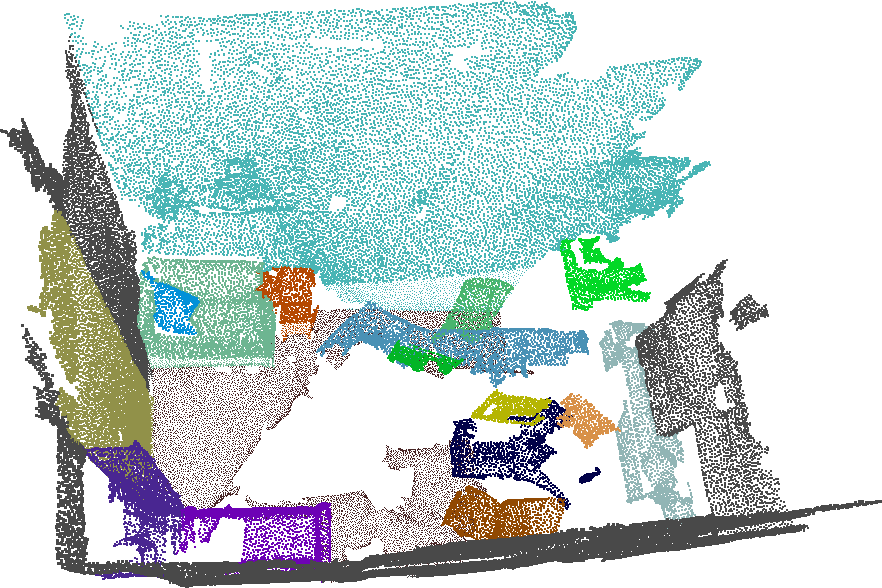}&
		\includegraphics[width=0.2\textwidth,height=0.13\textheight]{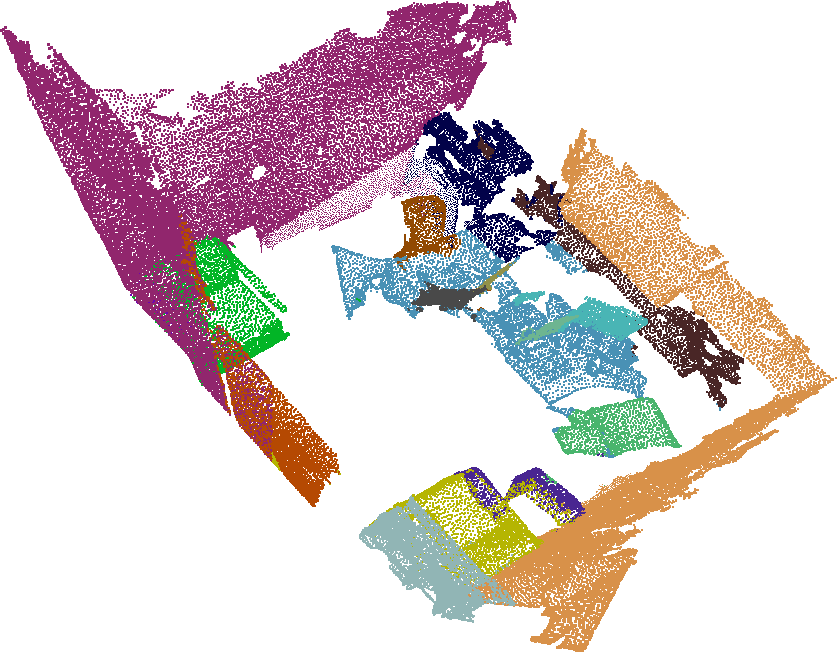}&
		\includegraphics[width=0.2\textwidth,height=0.13\textheight]{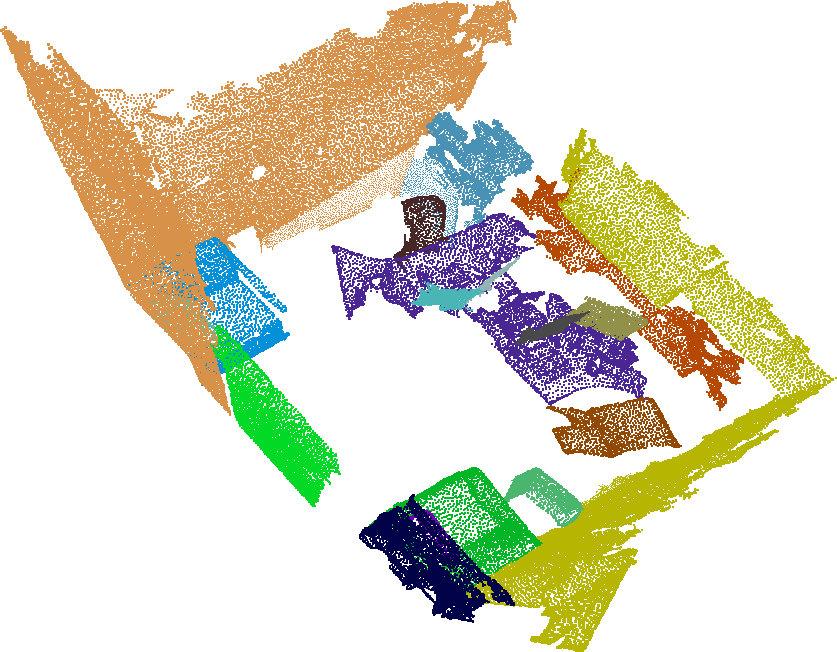}
		\\
		\includegraphics[width=0.2\textwidth,height=0.13\textheight]{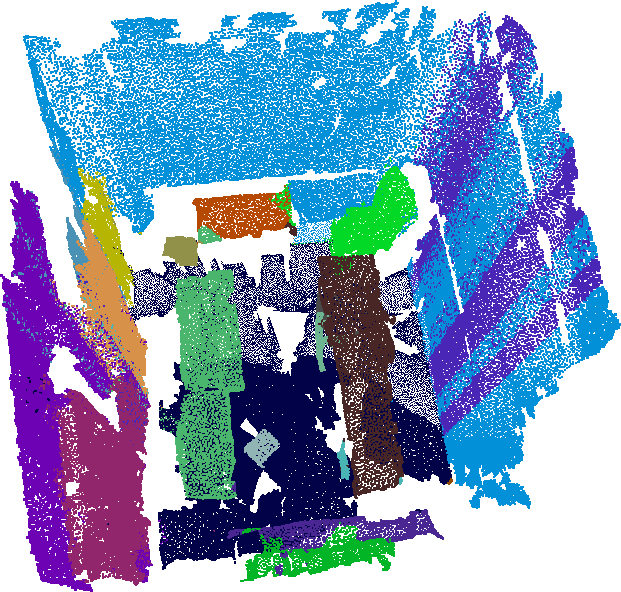}&
		\includegraphics[width=0.2\textwidth,height=0.13\textheight]{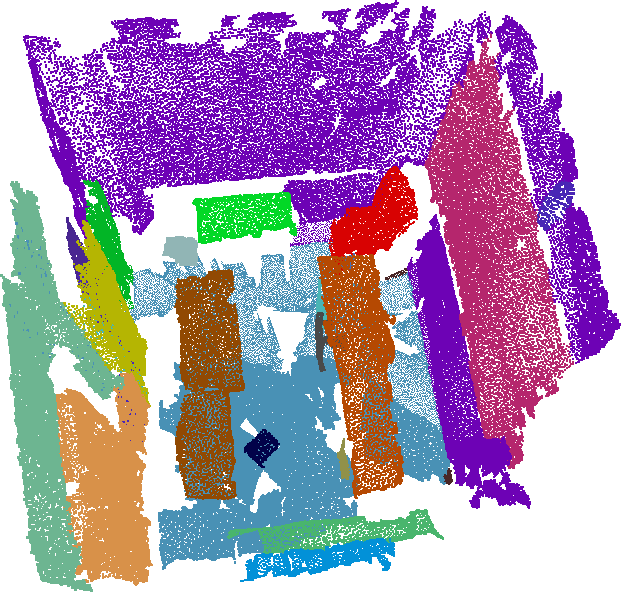}&
		\includegraphics[width=0.2\textwidth,height=0.13\textheight]{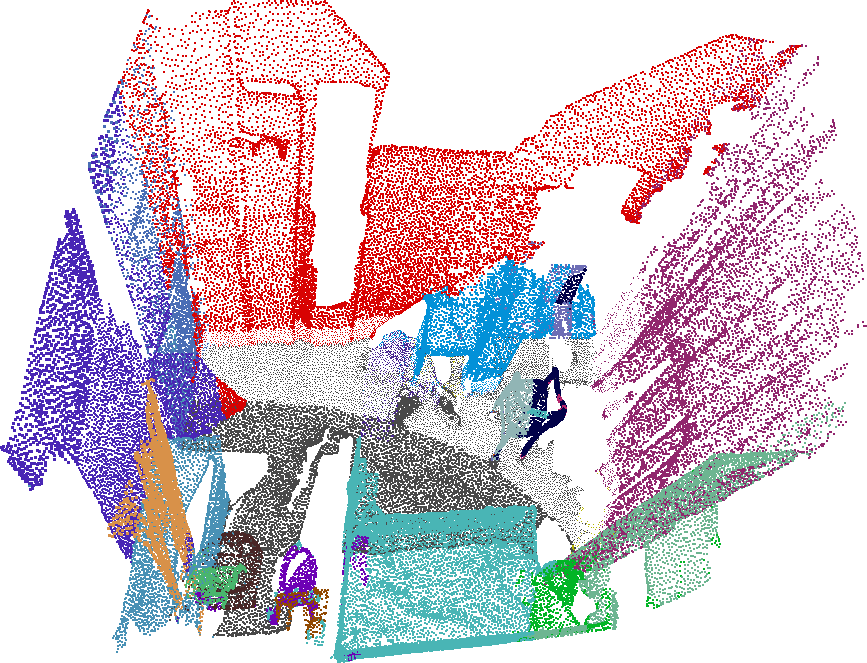}&
		\includegraphics[width=0.2\textwidth,height=0.13\textheight]{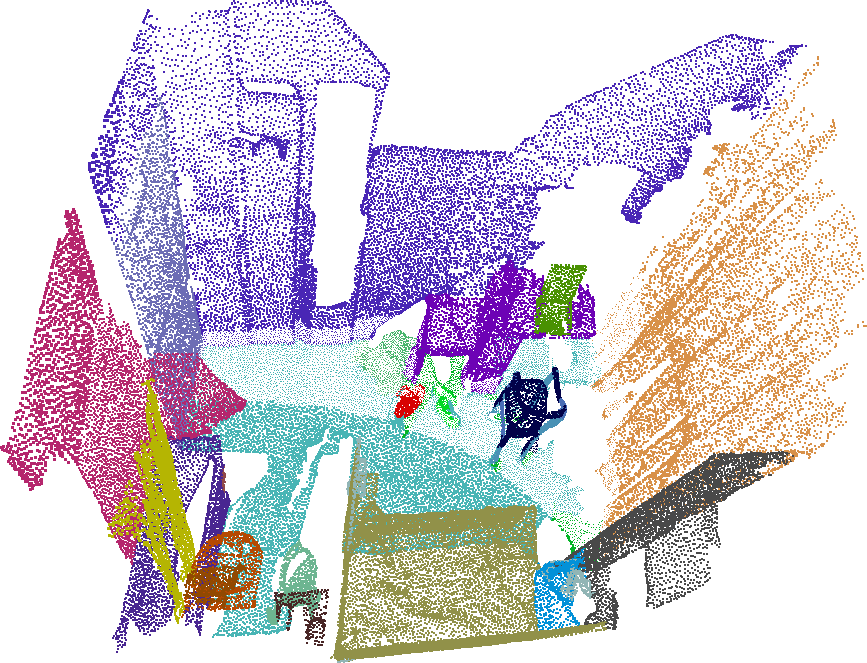}
		\\
		\includegraphics[width=0.2\textwidth,height=0.13\textheight]{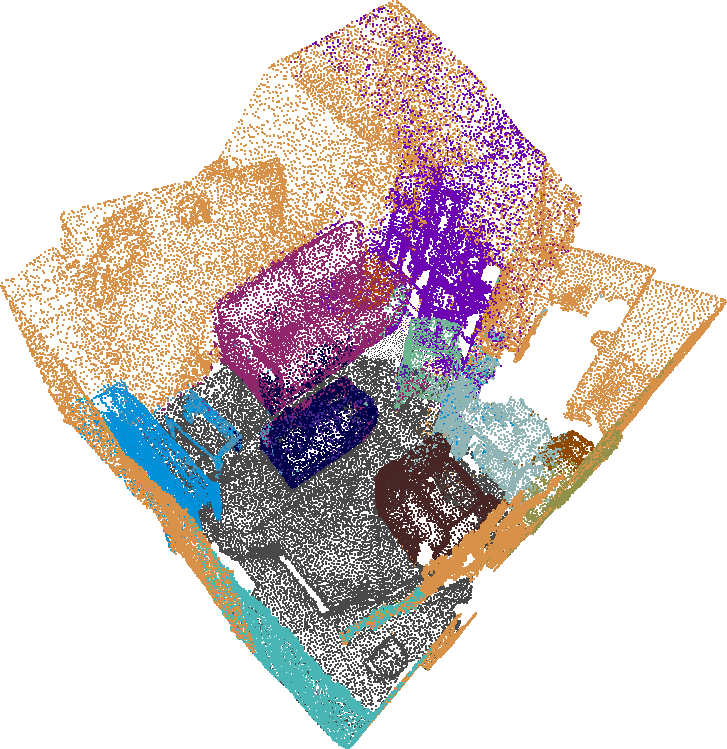}&
		\includegraphics[width=0.2\textwidth,height=0.13\textheight]{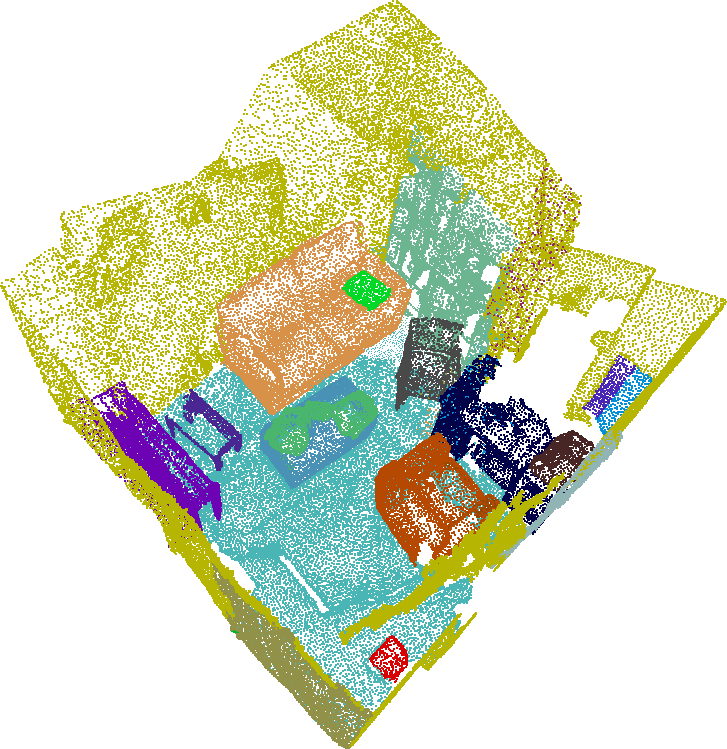}&
		\includegraphics[width=0.2\textwidth,height=0.13\textheight]{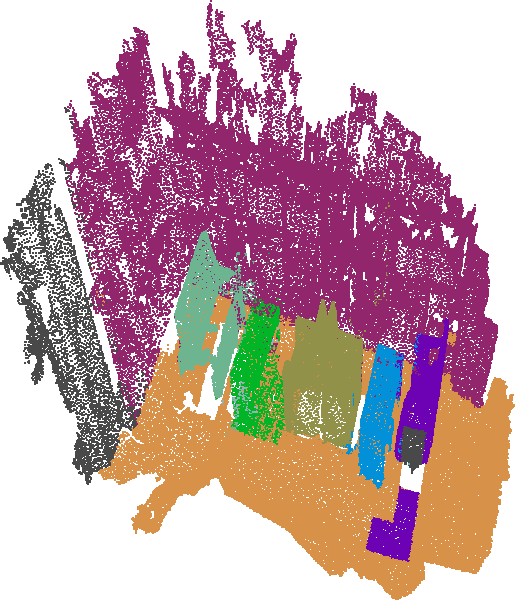}&
		\includegraphics[width=0.2\textwidth,height=0.13\textheight]{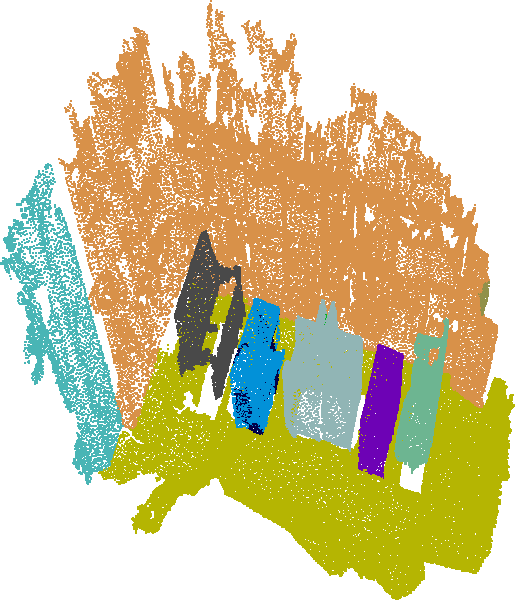}
		\\
		\includegraphics[width=0.2\textwidth,height=0.13\textheight]{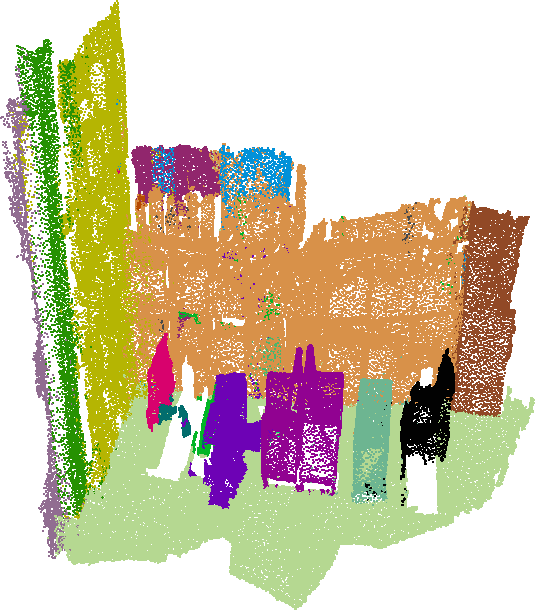}&
		\includegraphics[width=0.2\textwidth,height=0.13\textheight]{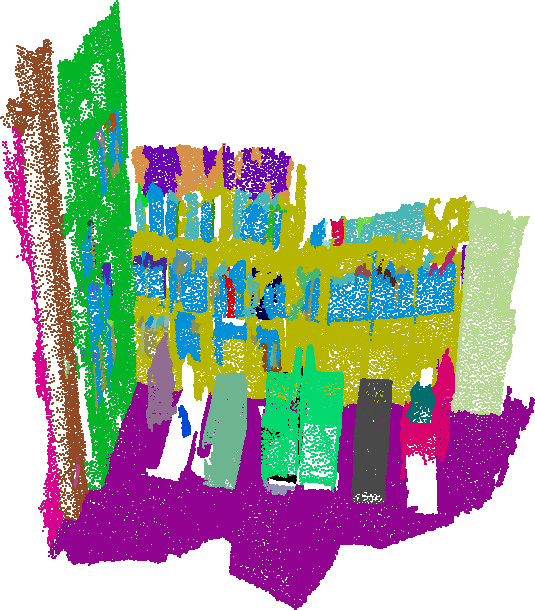}&
		\includegraphics[width=0.2\textwidth,height=0.13\textheight]{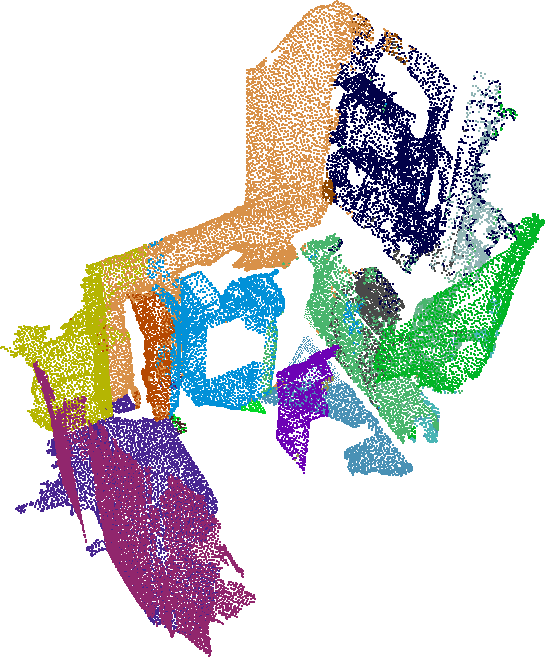}&
		\includegraphics[width=0.2\textwidth,height=0.13\textheight]{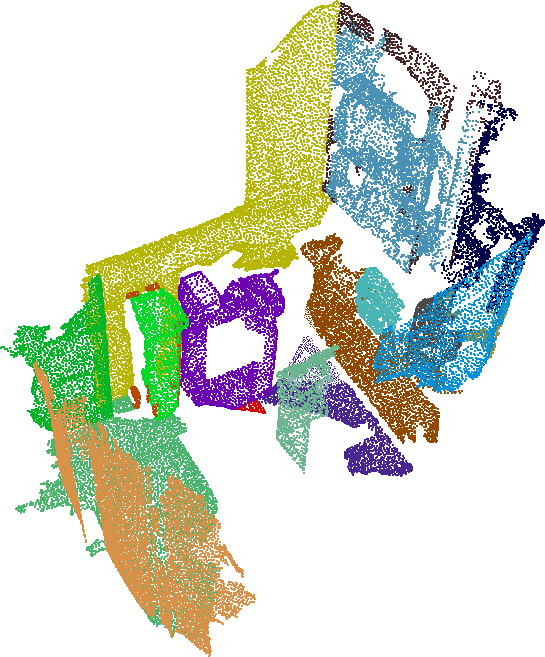}
		\\
		Prediction & Ground Truth & Prediction & Ground Truth
	\end{tabular}
	\caption{Instance segmentation results on ScanNet with SGPN. Different colors represent different instances. The colors of the same object in ground truth and prediction are not necessarily the same.}
	\label{fig:scannet}
\end{figure*}

We provide more experimental results on ScanNet~\cite{scannet}. This dataset contains $1513$ scanned and reconstructed indoor scenes. We use the official split with $1201$ scenes for training and $312$ for testing. Following the same \textit{BlockMerging} procedure, each scene is divided into $1.5m \times 1.5m$ blocks and each block is uniformly sampled into $4096$ points for training. All points in the block are used at test time. Each point is represented by a $9D$ vector (XYZ, RGB, and normalized location with respect to the room scene). PointNet++ is used as the baseline. The network architecture is:
\begin{flalign*}
&SA(1024, 0.1, [32, 32, 64]), \\
&SA(256, 0.2, [64, 64, 128]), \\
&SA(64, 0.4, [128, 128, 256]), \\
&SA(16, 0.8, [256, 256, 512]), \\
&FP(256, 256), \\
&FP(256, 256), \\
&FP(256, 128), \\
&FP(128, 128, 128, 128).
\end{flalign*}
And $F_{SIM}, F_{CF}, F_{SEM}$ are $1\times 1$ $conv$ layers with output channels $(128, 128, 128)$ respectively.
Table~\ref{table:scannet} illustrates the quantitative comparison results with Seg-Cluster. The metric is average precision (AP) with IoU threshold $0.25$.
Figure~\ref{fig:scannet} shows instance segmentation results on ScanNet.

